\documentclass[journal]{new-aiaa}
\usepackage[utf8]{inputenc}

\usepackage{graphicx}
\usepackage{amsmath}
\usepackage[version=4]{mhchem}
\usepackage{siunitx}
\usepackage{longtable,tabularx}
\usepackage{graphics}
\usepackage{epsfig} 
\usepackage{hhline}
\usepackage{amsmath, leftidx, mathtools, array} 
\usepackage{amssymb} 
\usepackage{multicol, lipsum, microtype}
\usepackage{array}
\usepackage{multirow}
\usepackage{tabu}
\usepackage{caption}
\usepackage{comment}
\usepackage{makecell}
\usepackage{tikz}
\usepackage{booktabs}
\usepackage{ctable}
\usepackage{tikz-3dplot}
\usepackage{pifont}
\usepackage[ruled,vlined,linesnumbered]{algorithm2e}
\usepackage[percent]{overpic}

\allowdisplaybreaks 

\usetikzlibrary{shapes}

\DeclareMathOperator*{\median}{median}
\DeclareMathOperator*{\maxcmd}{max}
\DeclareMathOperator*{\mincmd}{min}

\newcommand{\cmark}{\ding{51}}%
\newcommand{\xmark}{\ding{55}}%

\title{Momentum Model-based Minimal Parameter Identification \\ of a Space Robot}

\author{B. Naveen \footnote{Research Assistant, Department of Mechanical Engineering, N.H. 65 Nagour Road; naveen.13m@gmail.com} and Suril V. Shah \footnote{Assistant Professor, Department of Mechanical Engineering, N.H. 65 Nagour Road; surilshah@iitj.ac.in}}
\affil{Indian Institute of Technology, Jodhpur, Rajasthan, India, 342011}

\author{Arun K. Misra \footnote{ThomasWorkman Professor, Department of Mechanical Engineering, 817	Sherbrooke Street West; arun.misra@mcgill.ca. Fellow AIAA}}
\affil{McGill University, Montreal, Quebec H3A 0C3, Canada}

\begin{document}

\maketitle

\begin{abstract}
Accurate information of inertial parameters is critical to motion planning and control of space robots. Before the launch, only a rudimentary estimate of the inertial parameters is available from experiments and computer-aided design (CAD) models. After the launch, on-orbit operations substantially alter the value of inertial parameters. In this work, we propose a new momentum model-based method for identifying the minimal parameters of a space robot while on orbit. Minimal parameters are combinations of the inertial parameters of the links and uniquely define the momentum and dynamic models. Consequently, they are sufficient for motion planning and control of both the satellite and robotic arms mounted on it. The key to the proposed framework is the unique formulation of momentum model in the linear form of minimal parameters. Further, to estimate the minimal parameters, we propose a novel joint trajectory planning and optimization technique based on direction combinations of joints' velocity. The efficacy of the identification framework is demonstrated on a 12 degrees-of-freedom, spatial, dual-arm space robot. The methodology is developed for tree-type space robots, requires just the pose and twist data, and scalable with increasing number of joints.
\end{abstract}

\section*{Nomenclature}
{\renewcommand\arraystretch{1.0}
\noindent\begin{longtable*}{@{}l @{\quad=\quad} l@{}}
	$n$ & Number of 1-DoF joints in the space robot \\	
	$L_{i}$ & Link with index $i$ \\
	$J_{i}$ & Joint with index $i$ \\
	$\Sigma_{I}$ & Inertial frame of reference \\
	$\Sigma_{i}$ & Reference frame attached to $L_{i}$ based on modified  Denavit–Hartenberg (DH) notation \\
	$\Sigma_{i}^{(c)}$ & Reference frame attached to $L_{i}$ at its center of mass having the same orientation as $\Sigma_{i}$\\
	${\textbf{\textit{c}}}_{i} \in \mathbb{R}^{3}$ & Position vector from $\Sigma_{I}$ to $\Sigma_{i}^{(c)}$ \\
	${\textbf{\textit{r}}}_{i} \in \mathbb{R}^{3}$ & Position vector from $\Sigma_{I}$ to $\Sigma_{i}$ \\
	$\textbf{\textit{a}}_{i} \in \mathbb{R}^{3}$ & Position vector from $\Sigma_{i}$ to $\Sigma_{i}^{(c)}$ \\
	${\textbf{\textit{v}}}_{i}, \omega_{i} \in \mathbb{R}^{3}$ & Linear velocity and angular velocity of $\Sigma_{i}$ \\
	$\textbf{\textit{p}}_{i}, \textbf{\textit{l}}_{i} \in \mathbb{R}^{3}$ & Linear and angular momentum of $L_{i}$ respectively \\
	$\textbf{\textit{p}}, \textbf{\textit{l}} \in \mathbb{R}^{3} $ & System linear and angular momentum respectively \\
	$\textbf{\textit{K}}_{i} \in \mathbb{R}^{6 \times 10}$ & Link kinematic matrix of $L_{i}$ \\
	$\textbf{\textit{G}} \in \mathbb{R}^{6 \times 10(n+1)}$ & Global kinematic  matrix \\
	$\leftidx{^{i}}{\textbf{\textit{R}}}_{j} \in$ SO(3) & Rotation matrix from $\Sigma_{i}$ to $\Sigma_{j}$ \\
	$\textbf{\textit{R}}_{i}$ $\in$ SO(3) & Rotation matrix from $\Sigma_{I}$ to $\Sigma_{j}$ \\
	$m_{i} \in \mathbb{R}$ & Mass of $L_{i}$ \\
	$\leftidx{^{i}}{\textbf{\textit{a}}}_{i} \in \mathbb{R}^{3}$ & Position vector from $\Sigma_{i}$ to $\Sigma_{i}^{(c)}$ expressed in $\Sigma_{i}$ \\
	$\leftidx{^{i}}{\textbf{\textit{b}}}_{j} \in \mathbb{R}^{3}$ & Position vector from $\Sigma_{i}$ to $\Sigma_{j}$ expressed in $\Sigma_{i}$ \\
	$\rho_{j} \in \mathbb{I}^{+} \bigcup \text{ } \{0\} $ & Parent link index of $j^{\text{th}}$ link \\
	$\alpha_{j} $ & Joint twist between $\Sigma_{\rho_j}$ and $\Sigma_{j}$ \\
	$\leftidx{^{i}}{\textbf{\textit{I}}}_{i}^{(c)} \in \mathbb{R}^{3 \times 3}$  & Inertia matrix of $L_{i}$ around $\Sigma_{i}^{(c)}$ \\ 
	$\leftidx{^{i}}{\textbf{\textit{I}}}_{i} \in \mathbb{R}^{3 \times 3}$ & Inertia matrix of $L_{i}$ around $\Sigma_{i}$ \\
	$\mathbf{0}$, $\textbf{\textit{O}}$, $\textbf{\textit{E}}$ & Zero vector, Zero matrix, and Identity matrix respectively of compatible dimensions \\
	$\textbf{\textit{v}}^{(j)}$ & $j^{\text{th}}$ component of vector $\textbf{\textit{v}}$ \\
	${\pmb{\phi}}_{i} \in \mathbb{R}^{10 \times 1} $ & Link parameter vector of $L_{i}$ \\
	$\pmb{\phi} \in \mathbb{R}^{10(n+1) \times 1}$ & Standard parameter vector \\
	$t_{i}$ & $i^{\text{th}}$ time instant \\
	$\theta_{js}, \theta_{jd} \in \mathbb{R}$ & Start and range parameters of the $j^{\text{th}}$ joint \\
	$w \in \mathbb{R}^{+}$ & Weighing factor \\
	$\kappa(\cdot)$ & Condition number \\
	$N_{t}$ & Number of trajectories of rates (Six for base twist + $n$ for joints rates)\\
	$N_{m}$ & Number of instants of discretization of the trajectory \\
	$h_{ij}$ & Base/joint velocity from the $i^{\text{th}}$ trajectory at $j^{\text{th}}$ instant \\
	$\theta^{k}_{ji}, \theta^{k}_{jf}$ & Interval trajectory parameters, initial and final joint position respectively of the $j^{\text{th}}$ joint  in the $k^{\text{th}}$ interval \\
	$t$ & Time measured since the beginning of the $k^{\text{th}}$ interval \\
	$T_p$ & Time period of the interval	\\
	$q_{j}$ & Joint position \\
	$\epsilon(\cdot)$ & Root mean squared (RMS) error \\
	$q^{ }_{j, 0}$, $q^{(s)}_{j, i}$ and $q^{(c)}_{j, i}$ & Coefficients of fourier series for the $j^{\text{th}}$ joint \\
	$N_{h}$ & Number of harmonics in the fourier series\\
	$\omega_{f}$ & Angular frequency of the fourier series
\end{longtable*}
All the position and velocity vectors are expressed in $\Sigma_{I}$ unless mentioned.}

\section{Introduction}
\lettrine{O}{ver} the past two decades, space robots have gained prominence by performing on-orbit tasks, such as assembly, inspection, refueling, and docking \cite{flores2014review}. They are also a potential solution to remove space debris and harvest retired satellites \cite{flores2014review}. On-orbit tasks demand an accurate motion planning and control of a space robot, which in turn rely on the kinematic and dynamic models \cite{umetani1989resolved, papadopolous1991nature, nanos2015avoiding, james2016reactionless}. Unlike a fixed-base robot, the kinematics of a space robot is a function of the mass, the position of center of mass (CoM), and the inertia tensor of the links, evident from the Generalized Jacobian Matrix \cite{umetani1989resolved}. Consequently, to perform kinematic and dynamic analysis, motion planning, and control, the knowledge of inertial parameters is indispensable.

For a rigid link, there are ten inertial parameters, i.e., one with a mass, three with a position of CoM, and six with an inertia tensor. All of them can be identified for individual links before the system assembly \cite{peterson2004mass, gobbi2011method}. However, it demands a complicated experimental setup. Further, in a dismantled state, the links do not include cables for power transfer and communication and piping for fluid supply, which are quite heavy \cite{schwartz2004system}. Also, the CAD models provide estimates of inertial parameters which are erroneous by more than 10\% \cite{gobbi2011method}. Moreover, an inaccurate estimate of the inertial parameters adversely affects the motion planning and control \cite{james2015design}. Furthermore, while operating on-orbit, a change in the fuel quantity and payload can substantially alter the inertial parameters \cite{murotsu1994parameter, rackl2013parameter}. Neither the pre-assembly experiments nor the CAD models account for such an on-orbit change in the inertial parameters. A solution to address all the above-mentioned issues is a model-based identification of the inertial parameters of the robot while on orbit.

In this work, we focus on identifying the minimal parameters of a space robot based on the momentum model while on orbit. The minimal parameters define the dynamic model uniquely \cite{mayeda1990base, lin1995minimal}. They are a linear combination of the link parameters, which are the mass ($m_{i}$), the product of mass and position of CoM ($m_{i}\leftidx{^{i}}{\textbf{\textit{a}}}_{i}$) components, and the inertia tensor \big($\leftidx{^{i}}{\textbf{\textit{I}}}_{i}$\big) components of multiple links. Such a linear combination is due to constraining the relative motion of an adjacent pair of links with a joint \cite{mayeda1990base}, referred as kinematic constraints. The inertial parameters are redundant in defining the dynamic model; consequently, only the minimal parameters can be identified \cite{ayusawa2014identifiability}. Hence, our focus is on identifying the minimal parameters; however, by using the momentum model.

The central proposition of the presented work is to formulate the momentum model in the linear form of minimal parameters. The proposition is built on the evidence that the dynamic model can be formulated in the linear form of minimal parameters \cite{lin1995minimal, ayusawa2014identifiability, atkenson1986estimation}. 
Formulating the momentum model in terms of minimal parameters makes them sufficient not only for momentum model-based control but also for motion planning and dynamic model-based control. Estimating such widely applicable minimal parameters of the space robot, with no prior knowledge of inertial parameters of any of its links, fundamentally distinguishes the presented work from other momentum-based identification techniques \cite{murotsu1994parameter, yoshida2002inertia, ma2008orbit, christidi2017parameter,  nguyen2013adaptive}.

Momentum model defining the kinematics is characteristic of a space robot. Using such a characteristic feature for minimal parameter identification, based on the proposed linear formulation of momentum model, has the following advantages: 
\begin{itemize}
	\item Minimal parameters fully construct the momentum and dynamic models. Hence, they are sufficient for modeling, motion planning, and control.
	\item Momentum model consumes low noise data; thus, results in accurate estimates.
	\item Identification framework does not require thrusters and fuel resources. Instead, electrically driven actuators are used.
	\item Momentum model is computationally efficient in comparison to dynamic model to use with adaptive control techniques under no external force and torque.
	\item Linear formulation is also the key to design computationally efficient trajectories for identifying the minimal parameters.
\end{itemize}

Firstly, a momentum model requires only pose and twist while a dynamic model needs torque and acceleration as well. On the one hand, position and velocity are easy to measure, and their measurements have high signal-to-noise (SN) ratio. On the other hand, torque and acceleration measurements often have low SN ratio in space robots since their signal amplitude is low for safety reasons \cite{yoshida2002inertia, ma2008orbit}. As a result, parameter estimates based on the momentum model are proven to be substantially accurate than dynamic model based parameter estimates \cite{murotsu1994parameter, christidi2017parameter}. Next, the proposed framework uses only robotic arms and reaction wheels for parameter identification. They consume only electricity; thus, preserving the limited fuel resources. Finally, an exact linear model is a significant component of an adaptive controller. Momentum model has only six equations for a system with an arbitrary number of links and consumes lower amount of data than the dynamic model. Thus, it is computationally efficient for designing an adaptive controller. 

\textbf{\textit{Contributions and Paper Organization.}} The primary contribution of this work is a systematic procedure for linear formulation of the momentum model in terms of the minimal parameters. In the usual momentum formulation of a space robot \cite{xu2017orbit, dmitrov2006dynamics}, inertial parameters are indistinct because they are grouped with kinematic data (base pose and twist, and joint angle and velocity) and geometric parameters (link length, joint twist, and joint offset). However, in Sec. \ref{sec:3}, the inertial parameters are systematically separated from the kinematic data and grouped into link parameters. Further, we obtain the System Kinematic Matrix (SKM) and standard parameters for a generic space robot. SKM contains the kinematic data of all the links of the system and links parameters of all the links are referred to as standard parameters. SKM linearly transforms the standard parameter vector to system momentum.


In Sec. \ref{sec:4}, the kinematic constraints due to the joints are imposed onto the momentum model. The kinematic constraints regroup the standard parameters resulting in a momentum model in the linear form of minimal parameters. A recursive approach is provided to obtain the closed-form solution of minimal parameters for a generic tree-type space robot based on only the geometric parameters of the system.

In Sec. \ref{sec:5}, computation of exciting joint trajectories is presented. They render all the minimal parameters identifiable and facilitate in estimating the parameters which are minimally sensitive to noise. The whole system usually cannot be excited with a single motion \cite{venture2009numerical}, and it presents a major challenge in computing exciting trajectories. It is tackled with a novel joint trajectory planning and optimization technique for computing exciting trajectories, which forms another contribution of the work. The novelty lies in composing the joint trajectory with multiple intervals to excite all the minimal parameters of the system. Each interval adds new information about the system behavior while improving the number of identifiable minimal parameters. Moreover, the proposed approach is computationally efficient and scalable with increasing degrees-of-freedom (DoF).

Finally, in Sec. \ref{sec:6}, we apply the identification procedure on a 12-DoF, spatial, dual-arm space robot and estimate its minimal parameters with measurement noise in the kinematic data. Further, the performance of the estimated minimal parameters is evaluated by predicting the system's kinematics and dynamics. The proposed method uses reaction wheels with known inertial parameters to apply momentum onto the robot. We assume that the minimal parameters of the space robot do not change while executing the exciting trajectories. Further, the geometric parameters are usually precisely known \cite{lin1995minimal, ma2008orbit} and assumed to be deterministic in nature.

\section{Related work}
Solving the problem of parameter identification usually entails model formulation, experiment design, and parameter estimation. We discuss the related literature from the perspective of these tasks.

\subsection{Modeling}
Identification techniques based on both momentum \cite{murotsu1994parameter, yoshida2002inertia, ma2008orbit, nguyen2013adaptive, christidi2017parameter} and dynamic \cite{schwartz2004system, murotsu1994parameter, rackl2013parameter, ayusawa2014identifiability, lampariello2005modeling, sekhavat2009ukf, tanygin1997mass} models exist. Many works involved estimating the parameters of one of the links, with the knowledge of inertial parameters of rest of them \cite{murotsu1994parameter, rackl2013parameter, ma2008orbit, nguyen2013adaptive, sekhavat2009ukf}. However, frameworks which identify atleast some inertial parameters or a combination of inertial parameters of all the links is referred in this work as whole-system identification. 

In one of the initial attempts towards whole-system identification, angular momentum model was linearized around mass and inertia parameters using partial differentiation \cite{yoshida2002inertia}. Adapting such a linearization to all the inertial parameters provides only an approximate linear model because of nonlinearity in the CoM parameters. Moreover, the framework in \cite{yoshida2002inertia} estimates only some of the parameters of all the links while needing the knowledge of others. 

Recently, a fusion of dynamic and momentum model was performed to estimate the inertial parameters of all the links \cite{xu2017orbit}. However, the procedure needs firing the thrusters and requires high-noise acceleration measurements. A combination of inertial parameters of the space robot are estimated using only the angular momentum model in \cite{christidi2017parameter}. It is the work that comes closest to whole-system identification based only on momentum model. However, the estimated parameters are limited to model-based control of attitude only. Hence, problems related to the position and linear velocity control, motion planning and collision avoidance in task space cannot be solved.

In \cite{venture2009numerical}, minimal parameters were estimated by considering only the base link's dynamic model. Thus the need for joint torque and joint acceleration are eliminated. Even though the methodology applies to generic tree-type floating-base systems, it is suitable for systems like humanoids and humans. They have huge acceleration leading to higher SN ratio in their base link's acceleration measurements as compared to space robots. The base link is a satellite in space robots, which has a very low acceleration for safety reasons \cite{ma2008orbit}.

It is evident that the whole-system identification involving dynamic models require high-noise measurements and may need fuel resources. Using the momentum model only some parameters are estimated, which are insufficient for motion planning and control in task space.

\subsection{Experiment Design}
The designed trajectories which facilitate identifying the parameters of interest are usually called exciting trajectories. An approach widely used for fixed-base robots is adapted for space robots in \cite{christidi2017parameter, nguyen2013adaptive, sekhavat2009ukf} by minimizing the condition number of the regressor matrix. A well-conditioned regressor matrix minimizes the effect of noise on the parameter estimates \cite{gautier1992exciting}. Also, the experiment design has to ensure that the SN ratio of the measured data is high \cite{rackl2013parameter} so that the perturbation in its ground truth is minimum. Both condition number and SN ratio are utilized in the current work.

In \cite{sekhavat2009ukf}, using ideas from optimal control, exciting trajectories were designed within 8 minutes for a single link 6-DoF system. However, with increasing number of dimensions, optimal control strategies often consume hours of computation time \cite{james2016reactionless}. Furthermore, exciting all the minimal parameters with a single motion is difficult \cite{venture2009numerical}. The difficulty increases with increasing DoF of the robot. This problem was partially solved by running a few intuitively selected trajectories on the robot \cite{ayusawa2014identifiability}. We identify that designing a systematic methodology for computing exciting trajectories as another area for research and propose a solution to it in this work. In Table \ref{comp_analysis}, a comparison of various whole-system identification techniques is shown.

\begin{table}[h!]
	\caption{A comparative analysis of model-based frameworks for whole-system parameter identification of a space robot}
	\label{comp_analysis}
		\begin{center}
			\begin{tabular}{cccccc}
				\cmidrule{1-6}\morecmidrules\cmidrule{1-6}
				\multirow{2}{*}{Approach} & \multicolumn{5}{c}{Comparision criterion} \\
				& A & B & C & D & E \\
				\cmidrule{1-6}
				Ayusawa et al. \cite{ayusawa2014identifiability} & \cmark & \cmark & \cmark & \xmark & \xmark \\
				Yoshida et al \cite{yoshida2002inertia} & \cmark & \xmark & \cmark & \cmark & \xmark \\
				Christidi et al. \cite{christidi2017parameter} & \cmark & \xmark & \cmark & \cmark & \cmark \\
				Xu et al. \cite{xu2017orbit} & \cmark & \cmark & \xmark & \xmark & \xmark \\
				Proposed work & \cmark & \cmark & \cmark & \cmark & \cmark \\
				\cmidrule{1-6}\morecmidrules\cmidrule{1-6}
			\end{tabular}
		\end{center}
		\textit{Comparison criterion description:} \\
		A: Framework applies to generic tree-type robotic systems \\
		B: Estimated parameters are sufficient for motion planning and control \\
		C: Framework does not need components consuming fuel \\
		D: Model does not require acceleration or torque measurements \\
		E: Systematic computation of exciting trajectories for parameter estimation \\
\end{table}

\subsection{Estimation Methodology}
In \cite{xu2017orbit}, owing to an unprocessed form of the momentum model, which is nonlinear in terms of parameters, particle swarm optimization was implemented for parameter estimation. Formulations based on a processed form of momentum model used least squares or an altered version of it for parameter estimation \cite{murotsu1994parameter, yoshida2002inertia, ma2008orbit, christidi2017parameter, nguyen2013adaptive}. Least squares worked well because the noise affecting the data that is required by the momentum model is usually quite low. Despite the presence of gravity-gradient torque, momentum model with least squares estimation performed well, evident from the momentum reconstructed by the estimated parameters in \cite{yoshida2002inertia}.

\section{Momentum Model in Linear Form of Standard Parameters} \label{sec:3}
In the usual momentum model formulation of a space robot, the inertial parameters are grouped among themselves and also with the kinematic data and geometric parameters \cite{xu2017orbit, dmitrov2006dynamics}. Thus, it is hard to conclude the identifiability of inertial parameters using such a momentum formulation. In this section, we formulate the momentum model for a generic tree-type space robot in the linear form of standard parameters. We consider an individual link in space, group its inertial parameters, and separate the kinematic data into a Link Kinematic Matrix (LKM). LKM transforms the link parameter vector to the link's momentum. By assembling LKMs of all the links, the system kinematic matrix (SKM) is obtained which linearly transforms the standard parameters of the space robot to its momentum.

\begin{figure}[t!]
	\centering
	\includegraphics[width=0.5\linewidth, keepaspectratio]{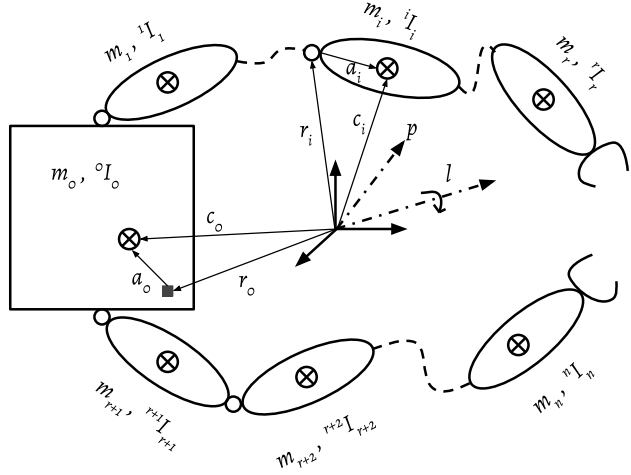}
	\caption{A dual-arm space robot with some of the notations. The sensor on the base link is at $\textbf{\textit{r}}_{0}$ in $\Sigma_{I}$} 
	\label{notation}
\end{figure}

A tree-type robot consisting of $n + 1$ rigid links connected by $n$ joints is considered. Without loss of generality, the joints are considered to be revolute with 1-DoF, which is often the case with space robots. One of the links is considered as a base link, which is a satellite or spacecraft in space robots. Further, body frames are assigned to the links based on the modified Denavit-Hartenberg (DH) convention. The base link is equipped with a sensor to measure its linear velocity and euler rates, shown in Figure \ref{notation}. To model the system, both linear and angular momentum are considered because angular momentum conservation does not necessarily imply linear momentum conservation and vice versa. 

\subsection{Linear Momentum of the $i^{th}$ Link}
It is well-known that the linear momentum is linear in terms of the link parameters \cite{ma2008orbit, nguyen2013adaptive} and for the $i^{\text{th}}$ link ($L_{i}$), it is as follows:
\begin{equation} \label{conv_link_lin_mtum}
\textbf{\textit{p}}_i = m_{i}\dot{\textbf{\textit{c}}}_{i}
\end{equation}

From the rigid body kinematics, the position and linear velocity of link CoM frame \big($\Sigma_{i}^{(c)}$\big) and link frame ($\Sigma_{i}$) are related as follows:
\begin{align} 
& \textbf{\textit{c}}_{i} = \textbf{\textit{r}}_{i} + \textbf{\textit{a}}_{i} \label{FK1} \\
& \dot{\textbf{\textit{c}}}_{i} = \textbf{\textit{v}}_{i} + \pmb{\omega}_{i} \times \textbf{\textit{a}}_{i} \label{FK2}
\end{align}
Substituting Eq. (\ref{FK1}) and Eq. (\ref{FK2}) in Eq. (\ref{conv_link_lin_mtum}) and separating the link parameters result in the link's linear momentum as:
\begin{equation} \label{lin_mtum_link_linear_form}
\textbf{\textit{p}}_i =  \textbf{\textit{v}}_{i}m_{i} +  ({\pmb{\tilde{\omega}}_{i}} \leftidx{^I}{\textbf{\textit{R}}_{i}})m_{i}\leftidx{^i}{\textbf{\textit{a}}_{i}}
\end{equation}
where, $\tilde{\pmb{\omega}}$ represents the skew-symmetrix matrix representation of $\pmb{\omega}$. \\
$\tilde{\pmb{\omega}} = 
 \begin{bmatrix}
	0 & -\omega_{z} & \omega_{y} \\
	\omega_{z} & 0 & -\omega_{x} \\
	-\omega_{y} & \omega_{x} & 0 \\
\end{bmatrix} $ 

\subsection{Angular Momentum of the $i^{th}$ Link}
For space robots, multiple works have used the nonlinear form of angular momentum for parameter estimation as in \cite{yoshida2002inertia, ma2008orbit, xu2017orbit}. However, with insights from the linear form of the dynamic model \cite{ayusawa2014identifiability, atkenson1986estimation}, we manipulate the angular momentum model on the similar lines to arrive at the linear form. The angular momentum of $L_{i}$ is as follows:

\begin{equation} \label{conv_ang_mtum}
\textbf{\textit{l}}_{i} =  \leftidx{^I}{\textbf{\textit{R}}_{i}}\leftidx{^i}{\textbf{\textit{I}}_{i}}^{(c)}\leftidx{^I}{\textbf{\textit{R}}_{i}}^{T}\pmb{\omega}_{i} + \textbf{\textit{c}}_{i}\times m_{i}\dot{\textbf{\textit{c}}}_{i}
\end{equation} 
Expanding Eq. (\ref{conv_ang_mtum}) using Eq. (\ref{FK1}) and Eq. (\ref{FK2}) and denoting the cross product with skew-symmetric notation for the kinematic data, we get:
\begin{equation} \label{ang_mtum_skew_symm_form}
\textbf{\textit{l}}_{i} =  \leftidx{^I}{\textbf{\textit{R}}_{i}}\leftidx{^i}{\textbf{\textit{I}}_{i}}^{(c)}\leftidx{^I}{\textbf{\textit{R}}_{i}}^{T} \pmb{\omega}_{i} + 
m_{i}\big({\tilde{\textbf{\textit{r}}}_{i}} \textbf{\textit{v}}_{i} + ({\tilde{\textbf{\textit{r}}}_{i}}  {\pmb{\tilde{\omega}}_{i}} - \tilde{\textbf{\textit{v}}}_{i}) \textbf{\textit{a}}_{i} + (\textbf{\textit{a}}_{i} \times \pmb{\omega}_{i} \times \textbf{\textit{a}}_{i})\big)
\end{equation}
Note that in Eq. (\ref{conv_ang_mtum}) and Eq. (\ref{ang_mtum_skew_symm_form}), the angular momentum is in nonlinear form of CoM parameters. However, it is linear in terms of mass and inertia. So, rewriting the nonlinear term $\textbf{\textit{a}}_{i} \times \pmb{\omega}_{i} \times \textbf{\textit{a}}_{i}$ as $\textbf{\textit{a}}_{i}^T\textbf{\textit{a}}_{i}\textbf{\textit{E}} - \textbf{\textit{a}}_{i}\textbf{\textit{a}}_{i}^{T}$ to get: 
\begin{flalign} \label{ang_mtum_nonlinear_last}
& \textbf{\textit{l}}_{i} =  \Big(\leftidx{^I}{\textbf{\textit{R}}_{i}}\leftidx{^i}{\textbf{\textit{I}}_{i}}^{(c)}\leftidx{^I}{\textbf{\textit{R}}_{i}}^{T} + m_{i}(\textbf{\textit{a}}_{i}^T\textbf{\textit{a}}_{i}\textbf{\textit{E}} - \textbf{\textit{a}}_{i}\textbf{\textit{a}}_{i}^{T})\Big)\pmb{\omega}_{i} + m_{i}\big({\tilde{\textbf{\textit{r}}}_{i}} \textbf{\textit{v}}_{i} + ({\tilde{\textbf{\textit{r}}}_{i}}  {\pmb{\tilde{\omega}}_{i}} - \tilde{\textbf{\textit{v}}}_{i}) \textbf{\textit{a}}_{i}\big) 
\end{flalign}
Substituting, $\textbf{\textit{a}}_{i} = \leftidx{^I}{\textbf{\textit{R}}_{i}}\leftidx{^i}{\textbf{\textit{a}}_{i}}$ in Eq. (\ref{ang_mtum_nonlinear_last}). As a result of the parallel axis theorem, nonlinear CoM terms are eliminated by grouping them with inertia around $\Sigma_{i}^{(c)}$ resulting in inertia around $\Sigma_{i}$ as follows:
\begin{equation} \label{penultimate_ang_mtum}
\textbf{\textit{l}}_{i} =  \big(\leftidx{^I}{\textbf{\textit{R}}_{i}}\leftidx{^i}{\textbf{\textit{I}}_{i}}\leftidx{^I}{\textbf{\textit{R}}_{i}}^{T}\big)\pmb{\omega}_{i} + m_{i}\big({\tilde{\textbf{\textit{r}}}_{i}} \textbf{\textit{v}}_{i} + ({\tilde{\textbf{\textit{r}}}_{i}}  {\pmb{\tilde{\omega}}_{i}} - \tilde{\textbf{\textit{v}}}_{i}) \leftidx{^I}{\textbf{\textit{R}}_{i}}\leftidx{^i}{\textbf{\textit{a}}_{i}}\big)
\end{equation}
where, $\leftidx{^i}{\textbf{\textit{I}}_{i}} = \leftidx{^i}{\textbf{\textit{I}}_{i}}^{(c)}+ m_{i}(\leftidx{^i}{\textbf{\textit{a}}}_{i}^T\leftidx{^i}{\textbf{\textit{a}}_{i}}\textbf{\textit{E}} - \leftidx{^i}{\textbf{\textit{a}}}_{i}\leftidx{^i}{\textbf{\textit{a}}}_{i}^{T})$ \\
Eq. (\ref{penultimate_ang_mtum}) is in the linear form of link parameters. Rewriting Eq. (\ref{penultimate_ang_mtum}) by separating the link parameters from the kinematic data as follows:
\begin{flalign} \label{ang_mtum_link_linear_form}
& \textbf{\textit{l}}_{i} =  \Big(\leftidx{^I}{\textbf{\textit{R}}_{i}}\big[(\leftidx{^I}{\textbf{\textit{R}}_{i}}^{T}\pmb{\omega}_{i}) \bullet\big]\Big)\big[\bullet\leftidx{^i}{\textbf{\textit{I}}_{i}}\big] + \big({\tilde{\textbf{\textit{r}}}_{i}} \textbf{\textit{v}}_{i}\big)m_{i} + \big(({\tilde{\textbf{\textit{r}}}_{i}}  {\pmb{\tilde{\omega}}_{i}} - \tilde{\textbf{\textit{v}}}_{i}) \leftidx{^I}{\textbf{\textit{R}}_{i}}\big)m_{i}\leftidx{^i}{\textbf{\textit{a}}_{i}} 
\end{flalign}
where, 
$[\bullet \textbf{\textit{I}}]$ = $\begin{bmatrix}
	I_{11} & I_{22} & I_{33} & I_{12} & I_{23} & I_{13}
\end{bmatrix}^{T}$, \quad 
$ [\pmb{\omega} \bullet]$ =
$ \begin{bmatrix}
\omega_{x} & 0 & 0 & 0 & \omega_{z} & \omega_{y} \\
0 & \omega_{y} & 0 & \omega_{z} & 0 & \omega_{x} \\
0 & 0 & \omega_{z} & \omega_{y} & \omega_{x} & 0 \\
\end{bmatrix}$ \\
Finally, Eq. (\ref{ang_mtum_link_linear_form}) is in the desirable linear form of the link parameters. 

\subsection{System's Total Momentum in Linear Form}
Total momentum of $L_{i}$ is obtained by appending the linear and angular momentum equations, i.e, Eq. (\ref{lin_mtum_link_linear_form}) and Eq. (\ref{ang_mtum_link_linear_form}) into a matrix-vector product form as follows:
\begin{equation}\label{link_kinematic_matrix}
\textbf{\textit{K}}_{i}{\pmb{\phi}}_{i} = \begin{bmatrix}
\textbf{\textit{p}}_{i} \\
\textbf{\textit{l}}_{i}
\end{bmatrix}
\end{equation}
where, 
\begin{flalign*} 
& \textbf{\textit{K}}_{i} = 
\begin{bmatrix}
\mathbf{\textit{O}} & \textbf{\textit{v}}_{i} & {\pmb{\tilde{\omega}}_{i}} \leftidx{^I}{\textbf{\textit{R}}_{i}} \\
\leftidx{^I}{\textbf{\textit{R}}_{i}}\big[(\leftidx{^I}{\textbf{\textit{R}}_{i}}^{T}\pmb{\omega}_{i}) \bullet\big] & {\tilde{\textbf{\textit{r}}}_{i}} \textbf{\textit{v}}_{i} & ({\tilde{\textbf{\textit{r}}}_{i}}  {\pmb{\tilde{\omega}}_{i}} - \tilde{\textbf{\textit{v}}}_{i}) \leftidx{^I}{\textbf{\textit{R}}_{i}}
\end{bmatrix} \in \mathbb{R}^{6 \times 10} & \\
& {\pmb{\phi}}_{i} = 
\begin{bmatrix}
\big[\bullet\leftidx{^i}{\textbf{I}_{i}}\big]^T &
m_{i} & 
m_{i}\leftidx{^i}{\textbf{a}}_{i}^{T}
\end{bmatrix}^T	\in \mathbb{R}^{10 \times 1} &
\end{flalign*}

The matrix $\textbf{\textit{K}}_{i}$ is the LKM, which consists of only the kinematic data of $L_{i}$. The momentum of the entire system is obtained by assembling the LKMs of all the links resulting in a System Kinematic Matrix (SKM) denoted by $\textbf{\textit{K}}$.
\begin{equation} \label{SKM}
\textbf{\textit{K}} \pmb{\phi}
=
\begin{bmatrix}
\textbf{\textit{p}} \\
\textbf{\textit{l}}
\end{bmatrix}
\end{equation}
where, 
\begin{flalign*}
& \textbf{\textit{K}} = 
\begin{bmatrix}
\textbf{\textit{K}}_{0} & \hdots & 		\textbf{\textit{K}}_{i} & \hdots & 
\textbf{\textit{K}}_{n}
\end{bmatrix} \in \mathbb{R}^{6 \times 10(n+1)} &
\\
& \begin{matrix}
\pmb{\phi}
\end{matrix} 
= 
\begin{bmatrix}
{\pmb{\phi}}_{0}^{T} &
\hdots &
{\pmb{\phi}}_{i}^{T} &
\hdots &
{\pmb{\phi}}_{n}^{T}
\end{bmatrix}^T \in \mathbb{R}^{10(n+1) \times 1} &
\end{flalign*}

The linear system of equations in Eq. (\ref{SKM}) is evaluated for kinematic data (base pose and twist, joint angle and velocity) at multiple instants and vertically appended together. The resultant is a global kinematic matrix (GKM) denoted by $\textbf{\textit{G}}$. 
\begin{equation} \label{GKM}
\textbf{\textit{G}}\pmb{\phi} = \textbf{\textit{m}}
\end{equation}
where,
\begin{flalign*}
& \textbf{\textit{G}} 
= 
\begin{bmatrix}
\textbf{\textit{K}}(t_1)^{T} & \hdots &	\textbf{\textit{K}}(t_{N_{m}})^{T} \\		
\end{bmatrix}^{T} \in \mathbb{R}^{6N_{m} \times 10(n+1)} &
\\
& \textbf{\textit{m}}
=
\begin{bmatrix}
\textbf{\textit{p}}({t_{1}})^{T} & \textbf{\textit{l}}({t_{1}})^{T} &
\hdots &
\textbf{\textit{p}}({t_{N_m}})^{T} & \textbf{\textit{l}}({t_{N_m}})^{T}
\end{bmatrix}^T \in \mathbb{R}^{6N_{m} \times 1} &
\end{flalign*}

Previously, it is shown in \cite{xu1993adaptive} that the inertia matrix is linear in terms of mass and inertia parameters of the links of the space robot. The inertia matrix can be used to construct the momentum model \cite{dmitrov2006dynamics}. We further show in Eq. (\ref{GKM}) that the momentum model is in linear form of standard parameters ($\pmb{\phi}$). Even with sufficient data, all the standard parameters are individually not identifiable because some of the columns of GKM are linearly dependent. Under the action of external force and torque on the space robot, the linear dependency in GKM is one of the following types: A) due to the kinematic constraints posed by the joints B) due to the choice of trajectory for constructing the GKM. Even for a fixed-base robot, linear dependencies occur due to same reasons with the matrix formulated using the linear form of dynamic model \cite{khosla1989categorization, gautier1992exciting}. Moreover, for a fixed-base robot, there are additional constraints due to its fixed base.

Type-A linear dependency is inherent to the GKM and exists irrespective of the robot, as long as 1-DoF joints connect the rigid links. Type-A linear dependency is not avoidable and has to be eliminated. Type-B is because of the chosen joint trajectories and the response of the base link to them. Type-B linear dependency is avoidable, unlike Type-A. In the next two sections, we overcome both the types of linear dependencies in the momentum model.

\section{Minimal Parameters and Regressor Matrix} \label{sec:4}
A joint connects two adjacent links in a robot and constraints their kinematics, i.e., their physical motion. In this section, we enforce these kinematic constraints onto the momentum model. Specifically, columns in GKM which are linearly dependent due to the kinematic constraints are identified. From the set of linearly dependent columns, we obtain a set of basis columns which are linearly independent. Rest of the columns are called redundant columns, which are a linear combination of the basis columns with the coefficients of linear combination. Further, Type A linear dependency is eliminated by deleting the redundant columns from the GKM resulting in the regressor matrix. Using the coefficients of linear combination, the standard parameters are regrouped, resulting in the minimal parameters. Grouping the standard parameters compensates for the deleted redundant columns in the momentum model. Hence, the momentum model stays intact, and it is in the minimal form. More properties of minimal parameters are discussed in \cite{mayeda1990base, lin1995minimal}.

On a case-by-case basis, the redundant columns and the linear dependency coefficients can be computed using matrix factorization techniques on the numerical GKM for a particular robot similar to \cite{lampariello2005modeling, gautier1991numerical}. We do not take such an approach because the identified linearly dependent columns offers no distinction between the columns of Type-A and Type-B. However, it is necessary to know the redundant columns due to Type-A exclusively, which should be eliminated. Exclusive knowledge of Type-A facilitates distinguishing Type-B linear dependency. Thus, the occurrence of Type-B linear dependency can be avoided.

We consider two adjacent links in a tree-type robot to find the redundant columns and their coefficients of linear combination with the basis columns. Using these coefficients, a recursive grouping of parameters is performed to obtain the minimal parameters of a tree-type space robot. Two links connected by a 1-DoF joint have a parent-child relationship. A parent is connected to the base link with a lower number of links than its child link. Each child has only one parent, and a parent can have multiple children. Base link has children, but it has no parent. Terminal links do not have children but have a parent. 

We use the following notation to represent the above mentioned relations between the links. For a child link with index $j, \text{ } j \in \mathbb{I}, \text{ } j > 0, \text{ } j \leq n + 1$, its parent is represented by $\rho_j$ ($\rho_j < i$). Further, for a parent with index $i$, the set of its children is represented by $\zeta_{i}$. The set of indices of all terminal links of a tree-type system is denoted by $T$ such that $\forall e \in T, \text{ } \zeta_{e} = \{\}$.

For the purpose of presentation, writing $\textbf{\textit{K}}_{i}$ from (\ref{link_kinematic_matrix}) as follows:
\begin{equation}
\textbf{\textit{K}}_{i}
=
\begin{bmatrix}
\textbf{\textit{K}}^{(1)}_{i} & \textbf{\textit{K}}^{(2)}_{i} & \textbf{\textit{K}}^{(3)}_{i} \\
\end{bmatrix}
\end{equation}
where,
\begin{flalign*}
& \textbf{\textit{K}}^{(1)}_{i} = 
\begin{bmatrix}
\mathbf{\textit{O}} & \big(\leftidx{^I}{\textbf{\textit{R}}_{i}}\big[(\leftidx{^I}{\textbf{\textit{R}}_{i}}^{T}\pmb{\omega}_{i})\bullet\big]\big)^{T}  \\
\end{bmatrix}^{T}  \in \mathbb{R}^{6 \times 6} & \\
& \textbf{\textit{K}}^{(2)}_{i} = 
\begin{bmatrix}
\textbf{\textit{v}}_{i}^{T} & ({\tilde{\textbf{\textit{r}}}_{i}} \textbf{\textit{v}}_{i})^{T} \\
\end{bmatrix}^{T} \in \mathbb{R}^{6 \times 1} & \\
& \textbf{\textit{K}}^{(3)}_{i} = 
\begin{bmatrix}
({\pmb{\tilde{\omega}}_{i}}\leftidx{^I}{\textbf{\textit{R}}_{i}})^{T} & (({\tilde{\textbf{\textit{r}}}_{i}}  {\pmb{\tilde{\omega}}_{i}} - \tilde{\textbf{\textit{v}}}_{i}) \leftidx{^I}{\textbf{\textit{R}}_{i}})^{T} 
\end{bmatrix}^{T} \in \mathbb{R}^{6 \times 3} &
\end{flalign*} 

Consider a parent-child pair $i, j \text{ such that } \rho_j = i$. The relation between the pose and twist of links $i, j$ due to a revolute joint is as follows:
\begin{flalign}
& \textbf{\textit{r}}_{j} = \textbf{\textit{r}}_{i} +\textbf{\textit{b}}_{j} \label{fk_position}\\
& \leftidx{^I}{\textbf{\textit{R}}_{j}} = \leftidx{^I}{\textbf{\textit{R}}_{i}}\leftidx{^{i}}{\textbf{\textit{R}}_{j}} \label{fk_orientation}\\
& \textbf{\textit{v}}_{j} = \textbf{\textit{v}}_{i} + \pmb{\omega}_{i} \times \textbf{\textit{b}}_{j} \label{fk_lin_vel}\\
& \pmb{\omega}_{j} = \pmb{\omega}_{i} + \leftidx{^I}{\textbf{\textit{R}}_{j}} \label{fk_ang_vel}
\begin{bmatrix}
0 & 0 & \dot{q}_{j} 
\end{bmatrix}^{T}
\end{flalign}
where $\textbf{\textit{b}}_{j} = \leftidx{^I}{\textbf{\textit{R}}}_{i}\leftidx{^i}{\textbf{\textit{b}}}_{j}$ and $\dot{q}_{j}$ is the joint velocity of the $j^{\text{th}}$ joint

Eq. (\ref{fk_position}) to Eq. (\ref{fk_ang_vel}) are substituted in LKM of child link, i.e., $\textbf{\textit{K}}_{j}$, and the linearly dependendent columns of $\textbf{\textit{K}}_{i}$ and $\textbf{\textit{K}}_{j}$ are identified using their symbolic forms. Consequently, the following redundant columns and their coefficients of linear combination are obtained as follows:
\begin{gather} 
\textbf{\textit{K}}^{(1)}_{j}\big<2\big> = \textbf{\textit{K}}^{(1)}_{i}\textbf{\textit{t}}_{1, j} -  \textbf{\textit{K}}^{(1)}_{j}\big<1\big> \label{LD_start}\\
\textbf{\textit{K}}^{(2)}_{j} = -\textbf{\textit{K}}^{(1)}_{i}\Big[\bullet \leftidx{^{i}}{\tilde{\textbf{\textit{b}}}}_{j}^2\Big] + \textbf{\textit{K}}^{(2)}_{i} + \textbf{\textit{K}}^{(3)}_{i}\leftidx{^{i}}{\textbf{\textit{b}}}_{j} \\
\textbf{\textit{K}}^{(3)}_{j}\big<3\big> = \textbf{\textit{K}}^{(1)}_{i}\textbf{\textit{t}}_{2, j} + \textbf{\textit{K}}^{(3)}_{i}\textbf{\textit{t}}_{3, j}
\label{LD_end}
\end{gather}
\begin{flalign*}
& \text{where, for a matrix } \textbf{\textit{A}}, \textbf{ \textit{A}}\big<k\big> \text{ represents } k^{th} \text{ column of } \textbf{\textit{A}} \\
& \textbf{\textit{t}}_{1, j} = 
\begin{bmatrix}
1, &
\text{C}_{\alpha_{j}}^2, &
\text{S}_{\alpha_{j}}^2, &
0, &
\text{C}_{\alpha_{j}}\text{S}_{\alpha_{j}}, &
0
\end{bmatrix}^{T} \\
& \textbf{\textit{t}}_{2, j} = 
\begin{bmatrix}
2\leftidx{^i}{\textbf{\textit{b}}_j}^{(3)}\text{C}_{\alpha_{j}} - 2\leftidx{^i}{\textbf{\textit{b}}_j}^{(2)}\text{S}_{\alpha_{j}}, &
2\leftidx{^i}{\textbf{\textit{b}}_j}^{(3)}\text{C}_{\alpha_{j}}, &
- 2\leftidx{^i}{\textbf{\textit{b}}_j}^{(2)}\text{S}_{\alpha_{j}},
&		 
\leftidx{^i}{\textbf{\textit{b}}_j}^{(1)}\text{S}_{\alpha_{j}}, &
\leftidx{^i}{\textbf{\textit{b}}_j}^{(3)}\text{S}_{\alpha_{j}} - 	\leftidx{^i}{\textbf{\textit{b}}_j}^{(2)}\text{C}_{\alpha_{j}}, &
- \leftidx{^i}{\textbf{\textit{b}}_j}^{(1)}\text{C}_{\alpha_{j}}
\end{bmatrix}^{T} \\
& \textbf{\textit{t}}_{3, j} = 
\begin{bmatrix}
0, &
-\text{S}_{\alpha_{j}}, &
\text{C}_{\alpha_{j}}
\end{bmatrix}^{T} \\
& \text{C}_{\alpha} = \text{cos}(\alpha), \quad \text{S}_{\alpha} = \text{sin}(\alpha) &
\end{flalign*}

The choice of redundant columns is not unique \cite{lin1995minimal}. However, we choose columns with the highest index, in line with the traditional literature \cite{gautier1990direct}. The redundant columns are $2^{\text{nd}}$, $7^{\text{th}}$, and $10^{\text{th}}$ columns of LKM of the child link and correspond to the link parameters, $\leftidx{^{i}}{\textbf{\textit{I}}}_{i}^{(yy)}$, $m_{i}$, and $m_{i}\leftidx{^{i}}{\textbf{\textit{a}}}_{i}^{(z)}$. The linear dependency coefficients are obtained for an arbitrary parent-child link pair connected by a 1-DoF revolute joint. The redundant columns and the coefficients of linear dependency obtained in Eq. (\ref{LD_start}) to Eq. (\ref{LD_end}) are in coherence with those obtained using the dynamic model in \cite{gautier1990direct}. However, in \cite{gautier1990direct} additional constraints due to the fixed base also exist. Also, it is clear that the obtained linear dependency coefficients are only a function of the geometric parameters. Hence, the linear dependency holds good between LKMs of every parent-child pair in a tree-type system at every instant, irrespective of the joint trajectory. 

Further, it can be inferred from Eq. (\ref{LD_start}) to Eq. (\ref{LD_end}) that a parent-child link pair creates redundant columns in the LKM of the child. A base link has no parent. As a result, columns of LKM of the base link are linearly independent due to kinematic constraints posed by joints. However, LKMs of the base link's children have redundant columns. Since a terminal link does not have children, it does not contribute to the formation of redundant columns in LKM of other links. However, terminal links have a parent in which they are always the child. Hence, the LKM of terminal links have redundant columns. Rest of the links become both parent and child to different links. Hence, LKM of such a link has redundant columns and creates redundant columns in the LKM of their child links. 

Based on the above-discussed logic, the links are divided into three categories, namely, link with no parent, links with no children, links with both parent and children. A recursive grouping of standard parameters corresponding to redundant columns is performed using the coefficients of linear dependency as follows:
\newline
For all links, $e \in T$ (set of indices of terminal links)
\begin{flalign} \label{rec_start}
\left.
\begin{aligned}
& \pmb{\tilde{\phi}}_{e} = \pmb{\phi}_{e} \\
& \pmb{\tilde{\phi}}_{e}^{(1)} = \pmb{\phi}_{e}^{(1)} - \pmb{\phi}_{e}^{(2)} \\
& \pmb{\phi}_{m, e} = D(\pmb{\tilde{\phi}}_{e}) \\
& \textbf{\textit{K}}_{m, e} = D(\textbf{\textit{K}}_{e})
\end{aligned}
\right\}
\end{flalign}
For all links, $i \notin T, \text{ } i \in \mathbb{I}, \text{ } i > 0, \text{ } i \leq n$
\begin{flalign}
\left.
\begin{aligned}
& \pmb{\tilde{\phi}}_{i} = \pmb{\phi}_{i} + \sum_{\forall j \in \zeta_{i}}\big(\pmb{\tilde{\phi}}_{j}^{(2)}\textbf{\textit{k}}_{1, j} + \pmb{\tilde{\phi}}_{j}^{(7)}\textbf{\textit{k}}_{2, j}	+ \pmb{\tilde{\phi}}_{j}^{(10)}\textbf{\textit{k}}_{3, j}\big) \\
& \pmb{\tilde{\phi}}_{i}^{(1)} = \pmb{\tilde{\phi}}_{i}^{(1)} - \pmb{\tilde{\phi}}_{i}^{(2)} \\
& \pmb{\phi}_{m, i} = D(\pmb{\tilde{\phi}}_{i}) \\
& \textbf{\textit{K}}_{m, i} = D(\textbf{\textit{K}}_{i})
\end{aligned}
\right\}
\end{flalign}
For the base link,
\begin{flalign} \label{rec_end}
& \pmb{\phi}_{m,0} = \pmb{\phi}_{0} + \sum_{\forall j \in \zeta_{0}}\big(\pmb{\tilde{\phi}}_{j}^{(2)}\textbf{\textit{k}}_{1, j} + \pmb{\tilde{\phi}}_{j}^{(7)}\textbf{\textit{k}}_{2, j}	+ \pmb{\tilde{\phi}}_{j}^{(10)}	\textbf{\textit{k}}_{3, j}\big)
\end{flalign}
where, $\pmb{\phi}_{m, 0} \in \mathbb{R}^{10 \times 1}$, $\pmb{\phi}_{m, i} \in \mathbb{R}^{7 \times 1} \text{ } \forall \text{ }i \in \mathbb{I}^{+}$, $\textbf{\textit{k}}_{1, j}, \textbf{\textit{k}}_{2, j}, \textbf{\textit{k}}_{3, j} \in \mathbb{R}^{10 \times 1}$ and
\begin{flalign*}
& 
\textbf{\textit{k}}_{1, j} 
=
\begin{bmatrix}
	\textbf{\textit{t}}^{T}_{1, j} & \textbf{\textit{0}}
\end{bmatrix}^T &
\\
& \textbf{\textit{k}}_{2, j} 
=
\begin{bmatrix}
	-\big[\bullet \leftidx{^{i}}{\tilde{\textbf{\textit{b}}}}_{j}^2\big]^T & 1 &
	\leftidx{^{i}}{\textbf{\textit{b}}}_{j}^T
\end{bmatrix}^{T} &
\\
& \textbf{\textit{k}}_{3, j} 
=
\begin{bmatrix}
	\textbf{\textit{t}}^{T}_{2, j} & 0 & \textbf{\textit{t}}^{T}_{3, j}
\end{bmatrix}^{T} &
\end{flalign*}

The function $D$ in Eq. (\ref{rec_start}) to Eq. (\ref{rec_end}) deletes the redundant columns from LKM, and the standard parameters corresponding to the redundant columns from link parameter vector. Note that the recursion process to obtain the closed-form solution of minimal parameters requires the knowledge of only the geometrical parameters.

Upon performing the recursive grouping process, we obtain the minimal SKM ($\textbf{\textit{K}}_{m}$) and the minimal parameter vector ($\pmb{\phi}_{m}$) which are:
\begin{flalign}
\textbf{\textit{K}}_{m} = 
\begin{bmatrix}
\textbf{\textit{K}}_{m, 0} & \hdots & \textbf{\textit{K}}_{m, n}
\end{bmatrix} \in  \mathbb{R}^{6 \times (10 + 7n)} \\
\pmb{\phi}_{m} = 
\begin{bmatrix}
\pmb{\phi}_{m, 0}^{T} & \hdots & \pmb{\phi}_{m, n}^{T}
\end{bmatrix}^{T}  \in  \mathbb{R}^{(10 + 7n) \times 1}
\end{flalign}

Previously, minimal parameters were obtained for a fixed-base robot using symbolic and numerical techniques based on the energy and dynamic models \cite{lin1995minimal, gautier1992exciting, gautier1990direct}. Further, the dynamics based formulation is also extended to space robots \cite{ayusawa2014identifiability}. Recently, a minimum set of parameters were obtained using only the angular dynamics and momentum model \cite{christidi2017parameter}. However, to our best knowledge, this is the first time minimal parameters are systematically derived based on both linear and angular momentum model of space robots. This reformulation of momentum model in terms of the minimal parameters forms the primary contribution of this work.

The momentum model in Eq. (\ref{GKM}) is rewritten using the regressor matrix ($\textbf{\textit{G}}_{m}$) and $\pmb{\phi}_{m} $ as  follows:
\begin{equation}\label{reg_mat}
\textbf{\textit{G}}_{m}\pmb{\phi}_{m} =
\textbf{\textit{m}}
\end{equation}
where, $
\textbf{\textit{G}}_{m} =
\begin{bmatrix}
\textbf{\textit{K}}_{m}(t_1)^{T} & \hdots &		\textbf{\textit{K}}_{m}(t_{N_{m}})^{T} \\		
\end{bmatrix}^{T} \in \mathbb{R}^{6N_{m} \times (10+7n)}.$
The momentum ($\textbf{\textit{m}}$) in Eq. (\ref{reg_mat}) is applied by the reaction wheels. Note that we have leveraged the conservation of momentum principle by conserving the summation of momentum of space robot and the momentum applied by the reaction wheels onto the space robot. The system of equations in Eq. (\ref{reg_mat}) with sufficient data points ($6N_{m} > 10+7n$) is overdetermined. A least squares estimate of the minimal parameter vector ($\pmb{\hat{\phi}}_{m}$) is:
\begin{equation} \label{param_estimate}
\pmb{\hat{\phi}}_{m} = (\textbf{\textit{G}}_{m}^{T}\textbf{\textit{G}}_{m})^{-1}\textbf{\textit{G}}_{m}^{T}\textbf{\textit{m}}
\end{equation}

Since the obtained minimal parameters fully construct the momentum model these parameters are sufficient for motion planning and momentum model-based control. Moreover, these minimal parameters are identical to the minimal parameters of floating-base robots based on the dynamic model in \cite{ayusawa2014identifiability}. Hence, estimating these parameters not only allow for momentum-based control but also for dynamic model-based control. In conclusion, they are sufficient for kinematics and dynamics analysis, motion planning, and control.

In this section, redundant columns due to Type-A linear dependency were identified and eliminated; thus, constructed the minimal parameter vector and regressor matrix. Next, we avoid the occurrence of  Type-B linearly dependency in the regressor matrix.

\section{Exciting Trajectories} \label{sec:5}
We present the other contribution of this work, which is the joint trajectory planning and optimization framework for computing exciting trajectories. First, an overview of the entire framework is provided, following which the methods and modules within the framework are discussed.

\subsection{Outline of the Proposed Framework}
\begin{figure}[t!]
	\centering
	\includegraphics[width=0.7\linewidth]{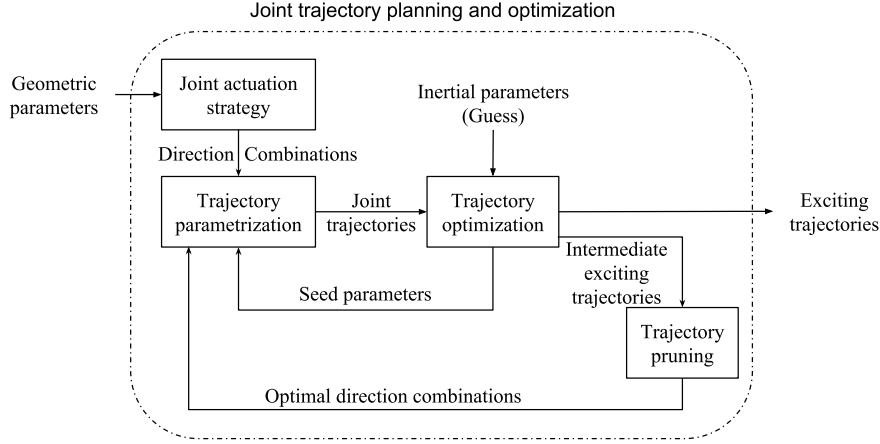}
	\caption{Block diagram of joint trajectory planning and exciting trajectories computation} 
	\label{block_diagram}
\end{figure}

The exciting trajectory computation framework majorly consists of joint actuation strategy, trajectory parametrization, optimization, and pruning modules, as shown in Figure \ref{block_diagram}. The geometric parameters of the robot and an initial guess of the inertial parameters are the inputs to the framework, and they can be obtained from the robot's CAD model.

Based on the geometric parameters of the robot, the joint actuation strategy decides the direction of joints' velocity during a time interval, which is encapsulated in direction combinations. Each interval executes a unique direction combination and consequently excites a subset of minimal parameters. The entire trajectory is composed of multiple intervals and thus excites all the minimal parameters, rendering them identifiable. The joint trajectory within a specific interval is parameterized with the interval trajectory parameters. In turn, the interval trajectory parameters of all the intervals of a particular joint are parameterized by the seed parameters. The seed parameters along with the knowledge of direction combinations parameterize the entire trajectory, hence, they are the design parameters of the trajectory optimization problem. Specifically, computing the exciting trajectories requires the following three steps:
\begin{itemize}
	\item Step 1: Parameterization of the joint trajectories with all the direction combinations and trajectory optimization, resulting in intermediate exciting trajectories
	\item Step 2: Pruning the intervals of the intermediate exciting trajectories, resulting in optimal direction combinations
	\item Step 3: Parameterization of the joint trajectories with only the optimal direction combinations, and trajectory optimization, resulting in optimal seed parameters and the exciting trajectories
\end{itemize}

In the block diagram in Figure \ref{block_diagram}, the inner loop containing the \textit{Joint trajectories} and \textit{Seed parameters} is executed in both Step 1 and Step 3 until the objective function converges to minima. The outer loop containing the \textit{Trajectory pruning} is executed only once, i.e., in Step 2. Next, we discuss the design of each module.

\subsection{Joints' Actuation Strategy} \label{direction_combinations}
We propose to excite the minimal parameters of a space robot using direction combinations of joint rates. A direction combination is a set of directions of the joint velocity of all the joints at a given instant. A non-zero joint velocity has two directions, either positive or negative. We fix the direction of the joint velocity of one joint constant throughout the entire duration. As a result, for a system with $n$ joints, there are $2^{n-1}$ direction combinations. Each direction combination is executed in a specific time interval. 

In \cite{rackl2012robot}, due to the inherent nature of B-splines, the exciting trajectory is composed of multiple intervals. However, in this work, the exciting trajectory is purposely composed with multiple intervals because each interval, composed with a unique direction combination, excites a subset of minimal parameters.

\subsection{Interval Trajectory Parametrization} \label{interval_param}
Here, parametrization of joint position and joint velocity in a time interval is discussed. The proposed interval trajectory parametrization facilitates achieving the joint actuation strategy.

A cycloidal function in time parameterizes the joint position. It requires just two parameters ($\theta^{k}_{jf}, \theta^{k}_{ji}$) to produce continuous and differentiable joint position, joint velocity, and joint acceleration. For $j^{\text{th}}$ joint during $k^{\text{th}}$ interval, $j, k \in \mathbb{I}$, $j, k > 0$, $j \leq n$,
and $k \leq 2^{n-1}$, 
the joint position and joint velocity are:
\begin{flalign}
& q^{k}_{j} = \theta^{k}_{ji} + \frac{(\theta^{k}_{jf} - \theta^{k}_{ji})}{T_{p}} \Bigg[t - \frac{T_{p}}{2\pi}\text{sin}\Bigg(\frac{2\pi t}{T_{p}}\Bigg)\Bigg] \label{cyc_pos}\\
& \dot{q}^{k}_{j} = \frac{(\theta^{k}_{jf} - \theta^{k}_{ji})}{T_{p}} \Bigg[1 - \text{cos}\Bigg(\frac{2\pi t}{T_{p}}\Bigg)\Bigg] \label{cyc_rate}
\end{flalign}

Along with having only two parameters to parameterize the trajectory in an interval, cycloidal parametrization also offers following advantages: 
\begin{itemize}
	\item Position function is monotonic in a given interval. The two parameters of the cycloidal function physically are the initial and final positions of the joint in an interval. Hence, by simply bounding those two parameters, the joint position can be contained within the desired limits.
	\item Since the position function is monotonic in a given interval, the joint velocity has a single direction in an interval. This behavior is extremely advantageous to execute the velocity combinations. By choosing a final position greater than initial position, a positive joint velocity is obtained. Similarly, for negative joint velocity, the final position should be lesser than the initial position.
	\item Further, analytical solutions exist for maxima and minima on velocity and acceleration and they result in linear constraints. Thus, constraining them within limits is simplified.
\end{itemize}

The joint trajectory is obtained by concatenating the interval trajectories. Further, the final position of the previous interval is the initial position of the current interval. Hence, for a single joint, $f$ intervals result in $f+1$ interval trajectory parameters. For a system with $n$ joints, there are $2^{n-1}$ intervals. Consequently, for $n$ joints, there are a total of $n(2^{n-1} + 1)$ interval trajectory parameters.

\subsection{Seed Parameters} \label{seed_param}
The interval trajectory parameters parameterize an interval trajectory. Here, seed parameters are introduced, which parameterize the interval trajectory parameters of all the intervals of a joint. The total number of interval trajectory parameters grows as $n(2^{n-1} + 1)$, which is undesirable for optimization and does not scale with increasing dimensions of the robot. Even with just six joints, the design space of optimization is of 198 dimensions. Hence, a small number of seed parameters are proposed to parameterize the $(2^{n-1} + 1)$ interval trajectory parameters of a joint.

For each joint, a seed consists of just two parameters, namely, start and range parameters. Start parameter is the initial position of the first interval. The range parameter is either added or subtracted to the initial position of an interval to get the final position. Addition or subtraction of range is decided by the direction requirement of the joint velocity in that interval. If a positive velocity is required, the range term is simply added, else subtracted. For $j^{\text{th}}$ joint during $k^{\text{th}}$ interval, the seed parameters construct the trajectory parameters as follows:
\begin{flalign}
& \theta_{ji}^{k} = 
\begin{cases}
\theta_{js}, & k = 1 \\
\theta_{jf}^{k - 1}, & k > 1 
\end{cases}	\label{seed_params1} \\
& \theta_{jf}^{k} = 
\begin{cases}
\theta_{jf}^{k - 1} + \theta_{jd}, & \dot{q}^{k}_{j} > 0 \\
\theta_{jf}^{k - 1} - \theta_{jd}, & \dot{q}^{k}_{j} < 0		
\end{cases} \label{seed_params2}
\end{flalign}
where, $\theta_{js}$ and $\theta_{jd}$ are the start and range parameters of the $j^{\text{th}}$ joint respectively. Since $\theta_{jd}$ is range, it is always greater than zero. A zero range leads to a constant joint position resulting in no joint motion. The interval trajectory parameters constructed in Eq. (\ref{seed_params1}) and Eq. (\ref{seed_params2}) are used to compute interval trajectories using Eq. (\ref{cyc_pos}) and Eq. (\ref{cyc_rate}). Using a seed, the design parameters of the optimization problem drastically reduce from $n(2^{n-1} + 1)$ to 2$n$. Sections V. \ref{interval_param} and V. \ref{seed_param} together comprise the \textit{Trajectory Parametrization} block in Figure \ref{block_diagram}.

\subsection{Trajectory Optimization} \label{trajectory_optimization}
The trajectory optimization problem is discussed in this subsection, which results in the optimal values of the seed parameters. The primary objective of exciting trajectories is to render all the minimal parameters identifiable. Specifically, the trajectory should not cause linear dependency (Type-B) in the columns of regressor matrix. Further, the sensor measurements are corrupted by noise. Hence, the secondary objective is to ensure that the estimates of the minimal parameters are minimally affected by noise. Exciting trajectories are computed by minimizing a cost function that fulfills the desired objectives.

\subsubsection{Objective Function}
We define an objective function involving the condition number of the regressor matrix and the magnitude of the base velocities and joint rates.
\begin{equation} \label{cost_fn}
\mincmd_{\theta_{js}, \theta_{jd}\text{ } \forall \text{ } j \in \text{1 to \textit{n}}} \quad \mathbf{C} = \kappa(\textbf{\textit{G}}_{m}) + w\frac{N_{t}N_{m}}{\sum_{i = 1 }^{N_{t}}\sum_{j = 1}^{N_{m}} h_{ij}^2} \\
\end{equation}
\textit{h} is the joint velocity or the component of base twist.
	
A low condition number ensures that the trajectory uniformly excites all the minimal parameters and results in the estimates which are relatively insensitive to noise. Moreover, the inverse of squares of base velocities and joints rates are added to the objective function to achieve trajectories measurements with high SN ratio. Since the measurement noise is independent of the measured value, higher velocity measurements correspond to a higher SN ratio. Measurements with low SN ratio do not represent the true behavior of the system, and the accuracy of estimates could be adversely affected. Both components of the objective function together help in realizing both the objectives of exciting trajectories.

\subsubsection{Simplified Joint Constraints}
The proposed interval trajectory parametrization with cycloidal trajectories and further parametrization of the interval trajectory parameters with the seed parameters not only reduces the number of design parameters of the optimization problem but also simplifies the joint constraints. We derive a simplified set of constraints, which result in trajectories adhering to joint angle, joint velocity, and joint acceleration limits.

The joint position can be simply constrained by bounding the interval trajectory parameters within the limits as follows:
\begin{equation}\label{position_constraint}
	\theta^{\text{min}}_{j} \leq \theta^{k}_{ji}, \theta^{k}_{jf} \leq \theta^{\text{max}}_{j} 
\end{equation}
Given the direction combinations, the interval trajectory parameters are computed from the seed parameters with only linear operations. Hence, the constraint in Eq. \ref{position_constraint} leads to only linear constraints on the design parameters of the optimization problem.

The maximum of the magnitude of $j^{th}$ joint velocity in $k^{th}$ interval is $\text{max}(||\dot{q}^{k}_{j}||) = 2(\theta_{jf}^{k} - \theta_{ji}^{k})/T_{p}$. However, for all $k$, $||(\theta_{jf}^{k} - \theta_{ji}^{k})|| = \theta_{jd}$, see Eq. (\ref{seed_params1}) and Eq. (\ref{seed_params2}). Hence, $\text{max}(||\dot{q}^{k}_{j}||) = 2\theta_{jd}/T_{p}$. To adhere to the joint velocity limit, $2\theta_{jd}/T_{p} \leq \text{min}(||\dot{\theta}_{j}^{\text{min}}||, ||\dot{\theta}_{j}^{\text{max}}||)$. On rearranging the equation, we get
\begin{equation}\label{velocity_limit}
	\theta_{jd} \leq (T_{p}/2)\text{min}(||\dot{\theta}_{j}^{\text{min}}||, ||\dot{\theta}_{j}^{\text{max}}||)
\end{equation}

Similarly, maximum of the magnitude of $j^{th}$ joint acceleration in $k^{th}$ interval can also be obtained, which is $\text{max}(||\ddot{q}^{k}_{j}||) = 2\pi\theta_{jd}/T_{p}^2$. To adhere to the joint acceleration limit, $2\pi\theta_{jd}/T_{p}^2 \leq \text{min}(||\ddot{\theta}_{j}^{min}||, ||\ddot{\theta}_{j}^{max}||)$. On rearranging the equation, we get
\begin{equation}\label{acceleration_limit}
	\theta_{jd} \leq (T_{p}^2/2\pi)\text{min}(||\ddot{\theta}_{j}^{\text{min}}||, ||\ddot{\theta}_{j}^{\text{max}}||)
\end{equation}
To satisfy both the Eq. (\ref{velocity_limit}) and Eq. (\ref{acceleration_limit}), 
\begin{equation}\label{range_param_upper_limit}
	\theta_{jd} \leq \text{min}\Big((T_{p}/2)\text{min}\big(||\dot{\theta}_{j}^{\text{min}}||, ||\dot{\theta}_{j}^{\text{max}}||\big),  (T_{p}^2/2\pi)\text{min}\big(||\ddot{\theta}_{j}^{\text{min}}||, ||\ddot{\theta}_{j}^{\text{max}}||\big)\Big)
\end{equation}

The right hand side of Eq. (\ref{range_param_upper_limit}) is simply a constant, represented by $\theta_{jd}^{\text{max}}$, which can be computed from the joint limits. Since, $\theta_{jd} > 0$, the bound constraint on $\theta_{jd}, \forall j \in \mathbb{I}, j > 0, j \leq n$ is as follows:
\begin{flalign}\label{range_constraint}
	0 < \theta_{jd} \leq \theta_{jd}^{\text{max}}
\end{flalign}

The proposed parametrization leads to only linear and bound constraints on the design parameters of the optimization problem for satisfying the joint limits. The 2$n$ design parameters are obtained by minimizing the objective function in Eq. (\ref{cost_fn}) with Eq. (\ref{position_constraint}) and Eq. (\ref{range_constraint})  as the constraints. Sequential Quadratic Programming (SQP) is used to solve the constrained optimization problem, which allows to include nonlinear constraints also.

\subsection{Pruning the Intermediate Exciting Trajectories} \label{trajectory_pruning}
Even though $2^{n-1}$ intervals add unique data about the system, not all of them are required to render the parameters identifiable and well-condition the regressor matrix. We present a methodology to prune such intervals. Each interval has a unique direction combination; pruning the intervals of intermediate exciting trajectories results in the optimal direction combinations. Moreover, with all the $2^{n-1}$ intervals, scaling the methodology to high-dimensional systems becomes infeasible because the time required for them to execute the exciting trajectories grows exponentially. 

To perform pruning (Step 2), we first compute the intermediate exciting trajectories, which consider all the direction combinations (Step 1). The process of pruning proceeds as shown in Algorithm \ref{prune_trajectory}. Regressor matrix ($\textbf{\textit{G}}_{m}$) is constructed using intermediate exciting trajectories, which is an input to pruning algorithm. Next, the regressor matrix constructed with the trajectory data of unpruned intervals is represented by ($\textbf{\textit{G}}_{m}^{o}$). Given the regressor matrix ($\textbf{\textit{G}}_{m}$) and the interval index ($i$), \textbf{getIntervalRM} retrives all the rows of $\textbf{\textit{G}}_{m}$ corresponding to the $i^{\text{th}}$ interval index represented by $\textbf{\textit{G}}_{m}^{i}$. First few intervals are unpruned until $\textbf{\textit{G}}_{m}^{o}$ has higher number of rows than columns ($n_{r} > n_{c}$). 

Next, every $\textbf{\textit{G}}_{m}^{i}$ is vertically concatenated to the $\textbf{\textit{G}}_{m}^{o}$, resulting in $\textbf{\textit{G}}_{m}^{t}$. The properties of $\textbf{\textit{G}}_{m}^{t}$ are compared with $\textbf{\textit{G}}_{m}^{o}$ to check if appending the regressor matrix of a particular interval has helped in achieving the objectives of computing exciting trajectories. Finally, $i^{\text{th}}$ interval from the rest of the intervals is unpruned if one of the following conditions is satisfied by $\textbf{\textit{G}}_{m}^{t}$:
\begin{enumerate}
	\item The number of significant singular values of $\textbf{\textit{G}}_{m}^{t}$, computed by the function \textbf{numSignificantSingularVals}, is more than that of $\textbf{\textit{G}}_{m}^{o}$, i.e., $s_{now} > s_{prev}$.
	\item The condition number of $\textbf{\textit{G}}_{m}^{t}$, computed by the function \textbf{condNum}, is less than that of $\textbf{\textit{G}}_{m}^{o}$, i.e., $cn_{now} < cn_{prev}$.
	\item The number of significant singular values of $\textbf{\textit{G}}_{m}^{t}$ are equal to $\textbf{\textit{G}}_{m}^{o}$ and occurs continuously for not more than $u_{max}$ times.
\end{enumerate}

\begin{algorithm}[t!]
	\caption{Prune intervals}
	\label{prune_trajectory}
	\DontPrintSemicolon
	\SetAlgoLined
	\SetKwInOut{Input}{Input\quad}
	\SetKwInOut{Output}{Output}
	\Input{$\textbf{\textit{G}}_{m}$, $n$, $u_{max}, \delta$}
	\Output{$\textbf{\textit{g}}$}
	$i, n_{r}, n_{c}, u  \leftarrow 0$ \;
	$\textbf{\textit{g}}, \textbf{\textit{G}}_{m}^{o} \leftarrow [\text{ }]$ \;
	\While{$n_{r} \leq n_{c}$}{
		$i \leftarrow i+1$ \;
		$\textbf{\textit{g}} \leftarrow [\textbf{\textit{g}}, i]$ \;
		$\textbf{\textit{G}}_{m}^{i} \leftarrow \textbf{getIntervalRM}(\textbf{\textit{G}}_{m}, i)$ \;
		$\textbf{\textit{G}}_{m}^{o} \leftarrow [(\textbf{\textit{G}}_{m}^{o})^{T}, (\textbf{\textit{G}}_{m}^{i})^{T}]^{T}$ \;
		$[n_{r}, n_{c}] \leftarrow \textbf{size}(\textbf{\textit{G}}_{m}^{o}) $ \;
	}
	$\textbf{\textit{G}}_{m}^{t} \leftarrow \textbf{\textit{G}}_{m}^{o} $\;
	$s_{prev} \leftarrow \textbf{numSignificantSingularVals}(\textbf{\textit{G}}_{m}^{o}, \delta)$ \;
	$cn_{prev} \leftarrow \textbf{condNum}(\textbf{\textit{G}}_{m}^{o})$ \;
	\While{$i < 2^{n - 1}$}{
		$i \leftarrow i+1$ \;
		$\textbf{\textit{G}}_{m}^{i} \leftarrow \textbf{getIntervalRM}(\textbf{\textit{G}}_{m}, i)$ \;
		$\textbf{\textit{G}}_{m}^{t} \leftarrow [(\textbf{\textit{G}}_{m}^{o})^{T}, (\textbf{\textit{G}}_{m}^{i})^{T}]^{T}$ \;		
		$s_{now} \leftarrow \textbf{numSignificantSingularVals}(\textbf{\textit{G}}_{m}^{t}, \delta)$ \;
		$cn_{now} \leftarrow \textbf{condNum}(\textbf{\textit{G}}_{m}^{t})$ \;
		\eIf{\Big(\em($s_{now} > s_{prev}$) \text{or} ($cn_{now} < cn_{prev}$) \text{or} \big(($s_{now} == s_{prev}$) \text{and} ($u \leq u_{max}$)\big)\Big)}{
			$\textbf{\textit{g}} \leftarrow [\textbf{\textit{g}}, i]$ \;
			$\textbf{\textit{G}}_{m}^{o} \leftarrow \textbf{\textit{G}}_{m}^{t}$ \;
			\eIf{\big(\em($s_{now} == s_{prev}$) \text{and} ($u \leq u_{max}$)\big)}{
				$u \leftarrow u + 1$ \;
			}
			{
				$u \leftarrow 0$ \;
			}
		}
		{
			$u \leftarrow 0$ \;
		}
	}
\end{algorithm}

Significant singular values are determined by the number of singular values whose magnitude is greater than a fraction ($\delta$) of the singular value with the highest magnitude. First condition means that unpruning a particular interval excites higher number of minimal parameters than without it. Second condition means that the system is well-conditioned with that particular interval than without it. Third condition is applied to reduce the effect of initial guess on the exciting trajectories. The index of every unpruned interval is stored in $\textbf{\textit{g}}$ whose elements are the indices of optimal direction combinations. Using the optimal direction combinations, the trajectory is parameterized and seed parameters are optimized, i.e. Step 3, also shown in Figure \ref{block_diagram}. The trajectory parameterized with the optimal direction combinations and the optimal seed parameters are the exciting trajectories to be executed on the system of interest.

\section{Results and Discussion} \label{sec:6}
In this section, we numerically apply the proposed identification framework on a spatial space robot. The exciting trajectories are computed, and they are executed by the space robot. The kinematic data of the robot is captured while performing the exciting trajectories, using which all the minimal parameters are estimated. Furthermore, the performance of the estimated minimal parameters is evaluated by performing inverse dynamics computations to calculate the error in the predicted base twist and joint torques in comparison to the ground truth.

\subsection{Space Robot Setup}
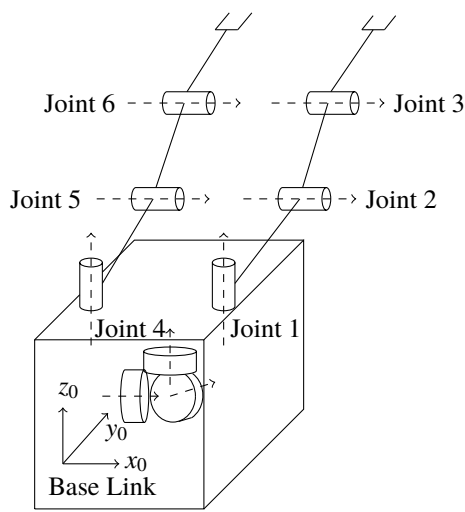
\begin{figure}[thpb]
	\centering
		\begin{tikzpicture}[scale = 0.75]
		\draw (0,0) -- (3,0) --(3,3) --(0,3) --(0,0);
		\draw (3, 3) -- (4.7678, 4.7678) -- (4.7678, 1.7678) -- (3, 0);	
		\draw (4.7678, 4.7678) -- (1.7678, 4.7678) -- (1.2, 4.2);
		\draw (0.8, 3.8) -- (0, 3);
		\node at (1.2, 0.4) {Base Link};
		\node (a) [cylinder, shape border rotate=90, draw, minimum height=0.7cm, minimum width=0.3cm, shape aspect = 0.5] at (3.35, 3.9){};
		\draw [dashed, ->] (3.35, 2.9) -- (3.35, 4.9);
		\node at (4.1, 3.2) {Joint 1};
		\draw (3.55, 4) -- (4.7, 5.5);
		\node (a) [cylinder, draw, minimum height=0.7cm, minimum width=0.3cm, shape aspect = 0.5] at (4.7, 5.5){};
		\node at (6.5, 5.5) {Joint 2};
		\draw [dashed, ->] (3.7, 5.5) -- (5.7, 5.5);
		\draw (4.75, 5.7) -- (5.2, 7.2);
		\node (a) [cylinder, draw, minimum height=0.7cm, minimum width=0.3cm, shape aspect = 0.5] at (5.2, 7.2){};
		\draw [dashed, ->] (4.2, 7.2) -- (6.2, 7.2);
		\node at (7, 7.2) {Joint 3};
		\draw (5.25, 7.4) -- (6, 8.5);
		\draw (6.2, 8.5) --  (5.8, 8.5);
		\draw (5.8, 8.5) -- (6, 8.75);
		\draw (6.2, 8.5) -- (6.4, 8.75);
		\node (a) [cylinder, shape border rotate=90, draw, minimum height=0.7cm, minimum width=0.3cm, shape aspect = 0.5] at (1, 3.9){};
		\draw [dashed, ->] (1, 2.9) -- (1, 4.9);
		\node at (1.7, 3.2) {Joint 4};
		\draw (1.2, 4) -- (2.1, 5.5);
		\node (a) [cylinder, draw, minimum height=0.7cm, minimum width=0.3cm, shape aspect = 0.5] at (2.1, 5.5){};
		\draw [dashed, ->] (1.1, 5.5) -- (3.1, 5.5);
		\node at (0.2, 5.5) {Joint 5};		
		\draw (2.15, 5.7) -- (2.65, 7.2);
		\node (a) [cylinder, draw, minimum height=0.7cm, minimum width=0.3cm, shape aspect = 0.5] at (2.65, 7.2){};
		\draw [dashed, ->] (1.65, 7.2) -- (3.65, 7.2);
		\node at (0.8, 7.2) {Joint 6};
		\draw (2.7, 7.4) -- (3.4, 8.5);
		\draw (3.2, 8.5) --  (3.6, 8.5);
		\draw (3.2, 8.5) -- (3.45, 8.8);
		\draw (3.6, 8.5) -- (3.85, 8.8);
		\draw [->] (0.5, 0.8) -- (1.5, 0.8);
		\draw [->] (0.5, 0.8) -- (0.5, 1.8);
		\draw [->] (0.5, 0.8) -- (1.3, 1.7);
		\node at (1.8, 0.8) {$x_{0}$};
		\node at (1.45, 1.35) {$y_{0}$};				
		\node at (0.6, 2.1) {$z_{0}$};
		\node (a) [cylinder, draw, minimum height=0.05cm, minimum width=0.7cm, shape aspect = 0.5] at (1.7, 2){};
		\draw [dashed, ->] (1.2, 2) -- (2.2, 2);
		\node (a) [cylinder, draw, minimum height=0.05cm, minimum width=0.7cm, shape aspect = 0.5, shape border rotate=90] at (2.4, 2.55){};
		\draw [dashed, ->] (2.4, 2.2) -- (2.4, 3.2);
		\draw (2.85,2) arc [x radius=0.4, y radius=0.46, start angle=0, end angle=57];
		\draw (2.85,2) arc [x radius=0.4, y radius=0.46, start angle=0, end angle=-235];
		\draw (2.8,2.4) arc [x radius=0.4, y radius=0.46, start angle=57, end angle=-60];
		\draw (2.55, 1.55) -- (2.78, 1.618);		
		\draw [dashed, ->] (2.4, 2) -- (3.2, 2.25);
	\end{tikzpicture}	
	\caption{The architecture of the 12-DoF, spatial, dual-arm space robot considered for the numerical study. The reaction wheels are mutually orthogonal with 3-DoF.} 
	\label{sim_model}
\end{figure}

With a growing interest in tasks like active debris removal and refurbishment of retired satellites, multi-arm space robots are expected to grow. Hence, we demonstrate the efficacy of the proposed framework by identifying the minimal parameters of a 12-DoF, spatial, dual-arm space robot, shown in Figure \ref{sim_model}. Each arm has three links connected by 1-DoF revolute joints. Two such arms contribute to 6-DoF. The base itself has 6-DoF resulting in a 12-DoF space robot. The true inertial parameters of the space robot are given in Table \ref{kin_dyn_params}. Terminal links, namely, $\textbf{\textit{L}}_{3}, \textbf{\textit{L}}_{6}$, hold a generic payload. Hence, their true inertial parameters are distributed along all the axes and higher than other links of the robotic arms. Further, the robot is equipped with three reaction wheels mounted on the base link along mutually orthogonal axes. However, the proposed identification framework applies to other arrangements of reaction wheels also. The closed-form solution of minimal parameters obtained using Eq. (\ref{rec_start}) to Eq. (\ref{rec_end}) for the robot shown in Figure \ref{sim_model} are available in the Appendix A. 1.

\begin{table}[t!]
	\caption{True inertial parameters of the robot and the offset inertial parameters. The offset inertial parameters are used to generate the exciting trajectories for minimal parameter identification}
	\label{kin_dyn_params}
	\begin{center}
				\begin{tabular}{*{9}{c}}
					\cmidrule{1-9}\morecmidrules\cmidrule{1-9}
					\thead{Parameters} & & $\textbf{\textit{L}}_{0}$ &  
					$\textbf{\textit{L}}_{1}$ &
					$\textbf{\textit{L}}_{2}$ &
					$\textbf{\textit{L}}_{3}$ &
					$\textbf{\textit{L}}_{4}$ &
					$\textbf{\textit{L}}_{5}$ &
					$\textbf{\textit{L}}_{6}$ \\
					\cline{1-9}
					\multirow{2}{*}{\thead{Mass \\ (Kg)}} & T & 2000 & 50 & 40 & 30 & 50 & 35 & 60 \\
					& O & 1500 & 30 & 20 & 50 & 40 & 25 & 30 \\
					\multirow{2}{*}{\thead{x-CoM \\ (m)}} & T & 0.2 & 0.6 & 0.4 & 0.7 & 0.55 & 0.45 & 0.6 \\
					& O & 0 & 0.4 & 0.7 & 0.5 & 0.4 & 0.3 & 0.5 \\
					\multirow{2}{*}{\thead{y-CoM \\ (m)}} & T & 0.3 & 0.05 & -0.45 & 0.4 & 0.04 & 0.05 & -0.5 \\
					& O & 0 & 0 & 0 & 0.2 & 0 & 0 & 0.2 \\
					\multirow{2}{*}{\thead{z-CoM \\ (m)}} & T & 0.4 & -0.07 & -0.05 & 0.3 & -0.04 & 0.05 & -0.35 \\
					& O & 0 & 0 & 0 & 0.1 & 0 & 0 & 0.2 \\
					\multirow{2}{*}{\thead{$I_{xx}^{c}$ \\ (Kg-$\text{m}^2$)}} & T & 1200 & 3.1 & 1.15 & 24.45 & 1.85 & 2.55 & 12.24 \\
					& O & 1000 & 2 & 1 & 14 & 1 & 1 & 6 \\
					\multirow{2}{*}{\thead{$I_{yy}^{c}$ \\ (Kg-$\text{m}^2$)}} & T & 1200 & 1.89 & 1.68 & 28.56 & 1.62 & 1.84 & 31.45 \\
					& O & 1000 & 1 & 1 & 20 & 2 & 1.5 & 20 \\
					\multirow{2}{*}{\thead{$I_{zz}^{c}$ \\ (Kg-$\text{m}^2$)}} & T & 1200 & 20.51 & 18.67 & 35.53 & 17.05 & 14.28 & 23.77 \\
					& O & 1000 & 10 & 28 & 20 & 9 & 19 & 35 \\
					\multirow{2}{*}{\thead{$I_{xy}^{c}$ \\ (Kg-$\text{m}^2$)}} & T  & 35.52 & 1.9 & 0.61 & 9.78 & 1.5 & 2.9 & 9.1 \\
					& O & 25 & 0.9 & 1.2 & 4.6 & 1 & 1.5 & 15 \\
					\multirow{2}{*}{\thead{$I_{yz}^{c}$ \\ (Kg-$\text{m}^2$)}} & T & 40.45 & 3.65 & 1.75 & 9.1 & 2.25 & 1.55 & 8.52 \\
					& O & 20.45 & 2.65 & 2.75 & 3.1 & 1.25 & 0.55 & 18.52 \\
					\multirow{2}{*}{\thead{$I_{zx}^{c}$ \\ (Kg-$\text{m}^2$)}} & T & 45.71 & 3.9 & 1.5 & 10.23 & 3.71 & 1.27 & 8.67 \\
					& O & 65.71 &  2.9 & 2.5 & 1.23 & 6.71 & 4.27 & 4.67\\
				\cmidrule{1-9}\morecmidrules\cmidrule{1-9}
				\end{tabular}
		\end{center}
		T and O represent true and offset inertial parameters respectively
\end{table}

\subsection{Exciting Trajectories}
Here, we discuss the details required to compute exciting trajectories, described in Figure \ref{block_diagram}, while providing insights into the proposed exciting trajectories computation framework.

\subsubsection{Direction Combinations} There are six joints in the space robot and three in the reactions wheels resulting in a total of nine joints. Hence, the number of direction combinations are $2^{9 - 1} = 256$, which is also the total number of intervals in the intermediate exciting trajectory. The robot executes each direction combination during a specific interval; we allot one second to each interval.

\subsubsection{Reaction Wheels Momentum} Each reaction wheel has predetermined seed parameters. The start and range parameters as 0 rad and 3$\pi$ rad respectively. They produce a maximum angular momentum of 18.8496 Kg-m$^2$/sec along its axis of rotation in its body frame. In case of a different reaction wheels arrangement, the entire framework remains the same. However, the other reaction wheels arrangement should execute the momentum trajectories computed using three mutually orthogonal wheels arrangement.

\subsubsection{Intermediate Exciting Trajectories} The objective function in Eq. (\ref{cost_fn}) is minimized using all the 256 direction combinations to obtain the intermediate exciting trajectories. Minimizing the cost requires kinematic data of the base link and the joints. Since base motion is a response to the joint trajectories according to the robot's momentum model, computing the cost requires an initial guess of the minimal parameters. 

For physical systems, an initial estimate is available from the CAD models of the system. However, to check the robustness of the methodology, we offset the true inertial parameters within $\pm$ 20\% - 120\% of their values, shown as offset parameters in Table \ref{kin_dyn_params}, and pass them as an initial guess to the exciting trajectory computation framework. Note that the guess of the terminal links and the base link are highly offset because on-orbit operations substantially modify the CAD model's estimates.

\subsubsection{Trajectory Pruning and Exciting Trajectories} Out of 256 intervals in the intermediate exciting trajectories, 189 of them are pruned using Algorithm \ref{prune_trajectory}. $u_{max}$ and $\delta$ in Algorithm \ref{prune_trajectory} are chosen as 5 and 1/300 respectively. Hence, the number of singular values whose magnitude is greater than 1/300 times the magnitude of the highest singular value is the number of significant singular values. As a result, 67 unique optimal direction combinations are obtained, which constitute the exciting trajectories. The result of pruning suggests that the proposed exciting trajectories computation framework is scalable with increasing DoF because only some of the $2^{n-1}$ direction combinations are required. The joint trajectories are parameterized with the 67 optimal direction combinations, and the seed parameters are optimized resulting in exciting trajectories. The optimal seed parameters of the exciting trajectories are tabulated in Table \ref{opti_tr_params}.

\subsubsection{Analysis of the Exciting Trajectories Computation Framework}
\begin{figure}
	\centering
	\includegraphics[width=0.5\linewidth, keepaspectratio]{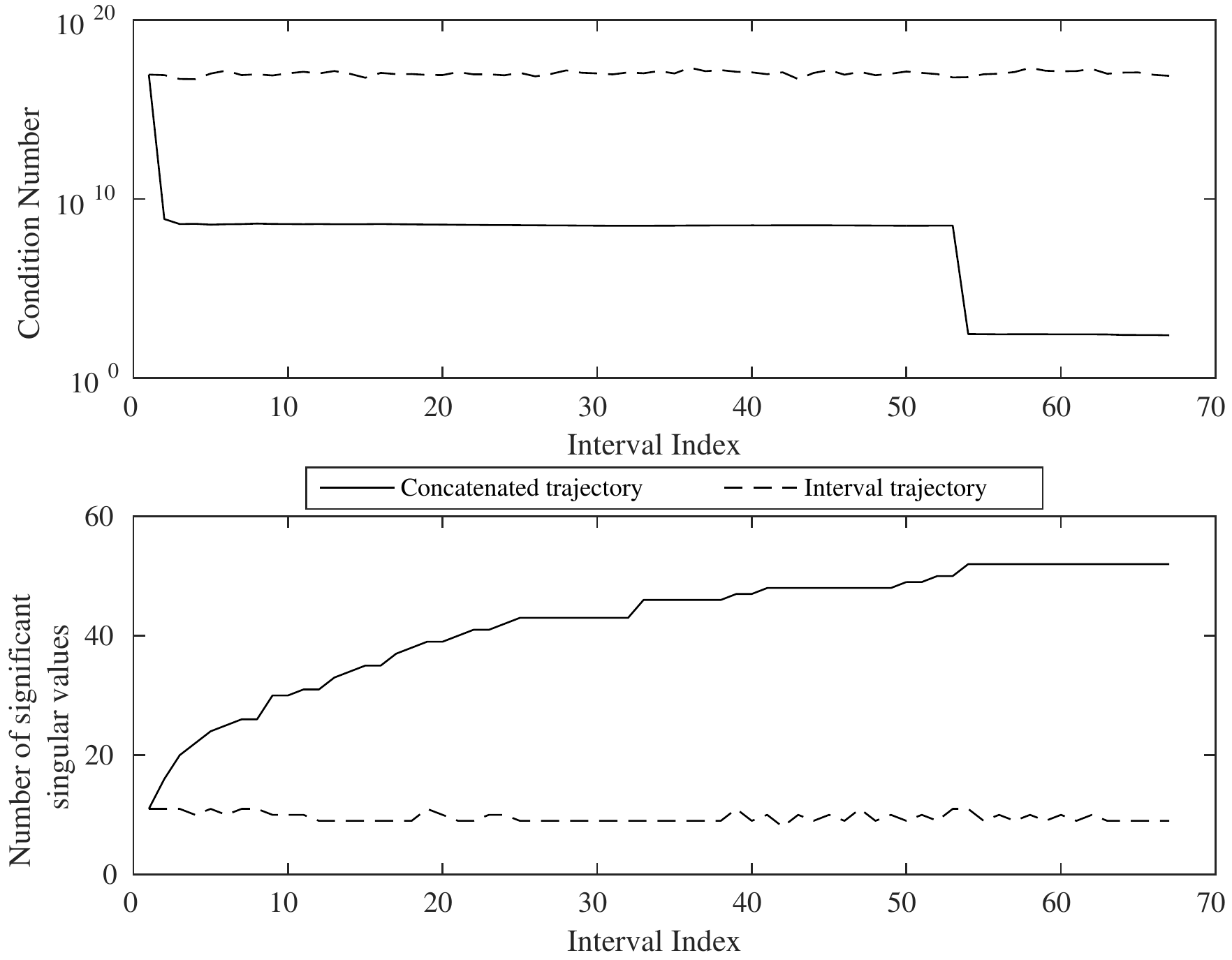}
	\caption{a) Condition number and b) Number of significant singular values of the regressor matrix obtained using exciting trajectories executed on the true system} 
	\label{cn_vs_interval}
\end{figure}

For analysis, using the exciting trajectories, the condition number of the regressor matrix and the number of significant singular values are computed for the true system, shown in Figure \ref{cn_vs_interval}. These computations are made for two kinds of trajectories, a) interval trajectory, which consists of one particular direction combination and b) concatenated trajectory, which consists of several interval trajectories appended together up to the interval index.

Each interval trajectory excites only some minimal parameters of the robot up to the desired level (in this case, 8 to 11 out of 52). However, different interval trajectories excite a different subset of minimal parameters. Consequently, all the optimal interval trajectories together excite all the minimal parameters, apparent from Figure \ref{cn_vs_interval}. Further, the condition number of the regressor matrix constructed with a single interval is extremely high ({\raise.17ex\hbox{$\scriptstyle\sim$}} 10$^{18}$) because it excites only some minimal parameters. However, optimal intervals and the optimal seed parameters have brought a significant reduction in the condition number to 248.8. The proposed concept of direction combinations resulted in interval trajectories. The interval trajectories together excited all the minimal parameters. Thus, the minimal parameters are rendered identifiable while also resulting in a low condition number of the regressor matrix.

Further, the number of seed parameters are only 2$n$ with the proposed technique, 5.5 times lower than the trajectory parameters of commonly used 5-term fourier series. Such a reduction in the number of design parameter for optimization makes it highly desirable to generate exciting trajectories even for high-dimensional space robots. Moreover, even after generating exciting trajectories with hugely offset inertial parameters, the trajectories excited all the minimal parameters and provided a regressor matrix with a low condition number. This indicates that the proposed framework for computing exciting trajectories is robust to a bad initial guess of inertial parameters.

In Figure \ref{cn_vs_interval}, the condition number takes a massive drop twice. The first drop occurs between the 1$^\text{st}$ and 3$^\text{rd}$ intervals. The number of significant singular values because of first interval is 11. Upon appeding the 2$^\text{nd}$ and 3$^\text{rd}$ intervals, the number of significant singular values improve to 20. Owing to a sudden improvement in the number of significant singular values led to a massive drop in the condition number. 	Such an improvement is possible initially because only 11 of the 52 minimal parameters are excited by the 1st interval. The chances of exciting new minimal parameters by the subsequent interval is high. The second drop takes place when number of significant singular values reach 52, which is equal to the number of minimal parameters. Until then, at least one minimal parameter was not sufficiently excited by the trajectory resulting in an ill-conditioned regressor matrix.

\begin{table}
	\caption{Seed parameters of exciting joint trajectories obtained with offset inertial parameters as initial guess to the trajectory optimizer}
	\label{opti_tr_params}
		\begin{center}
				\begin{tabular}{ccccccc}
					\cmidrule{1-7}\morecmidrules\cmidrule{1-7}
					\multirow{3}{*}{\thead{Seed \\ parameters \\ (deg)}} & \multicolumn{6}{c}{Joint index} \\
					& \multirow{2}{*}{1} & \multirow{2}{*}{2} & \multirow{2}{*}{3} & \multirow{2}{*}{4} & \multirow{2}{*}{5} & \multirow{2}{*}{6} \\ 
					&&&&&& \\
					\cmidrule{1-7}
					\thead{Start ($\theta_{s}$)} & 40.661 & -109.403 & -78.782  &  125.089  & -110  & 51.228  \\
					\thead{Range ($\theta_{d}$)} & 0.524 & 2.51 & 2.51 & 2.513  & 2.036 & 2.5  \\
					\cmidrule{1-7}\morecmidrules\cmidrule{1-7}
				\end{tabular}
	\end{center}
\end{table}

\subsection{Parameter Estimation}
In this subsection, the exciting trajectories are executed by the space robot in ReDySim \cite{shah2012recursive}, and the kinematic data is captured. Using the captured kinematic data, the regressor matrix is constructed and the minimal parameters are estimated using Eq. (\ref{param_estimate}). 
It may be noted that, unlike the existing whole-system momentum-based identification frameworks, we do not estimate only a few parameters of all the links. We estimate even the parameters including the CoM terms along the y-axis and z-axis and product of inertia of the links, which are ignored by the existing works.

To check the validity of the proposed identification framework, the regressor matrix is constructed using the kinematic data that is free of noise, and the minimal parameters are estimated. Without noise, as expected, the estimates match exactly with the true values, tabulated in Table \ref{min_params}.

For the simulated data to mimic the real data, zero-mean gaussian white noise is added to the joint rates and base velocities, as in \cite{ma2008orbit}. The standard deviation in measurement noise on linear and angular velocity components of the base and joint rates are 50 $\mu$m/sec, 80 $\mu$rad/sec, and 50 $\mu$rad/sec respectively. The regressor matrix is constructed with the noisy kinematic data and the minimal parameters are estimated, shown under Case 1 in Table \ref{min_params}. To understand the accuracy of the estimates, the median and maximum of the relative error of the estimated parameter estimates wrt. their true values are computed, and they are tabulated in Table \ref{param_error}. 
\begin{flalign}\label{avg_max_error}
&\epsilon_{median} = \median_{i = 1}^{52} \Bigg| \frac{\pmb{\phi}_{m}^{(i)} - \hat{\pmb{\phi}}_{m}^{(i)}}{\pmb{\phi}_{m}^{(i)}}  \Bigg| \\
& \epsilon_{max} = \maxcmd_{i = 1}^{52} \Bigg| \frac{\pmb{\phi}_{m}^{(i)} - \hat{\pmb{\phi}}_{m}^{(i)}}{\pmb{\phi}_{m}^{(i)}}  \Bigg|
\end{flalign}

The median relative error is as low as 2\%, and the maximum relative error is 174\%. Most of the estimated parameters are quite accurate apart from 10-11 parameters which have a relative error more than 10\%. Such a relative error occurred with parameters having comparatively low numeric values because they have a low contribution towards the system behavior; hence, noise easily distorts their estimates. These are the minimal parameters containing the product of inertia terms of the links of the robotic arm. However, terms containing the product of inertia of either the base link or the end effector link are accurately estimated because their numeric value is comparatively high. Even with the obtained relative error, in the next subsection, we show that these parameters predict the future states of the space robot accurately.

\begin{center}
	\begin{longtable}{lcccc}
		\caption{True and estimated value of minimal parameters of 12-DoF space robot with different joint trajectories}
		\label{min_params} \\
		
		\cmidrule{1-5}\morecmidrules\cmidrule{1-5}
		\multirow{2}{*}{\thead{\quad Minimal \\ \quad Parameter}} & \multirow{2}{*}{\thead{True \\ Value}} & \multicolumn{3}{ c }{\thead{Estimated Value}} \\
		& & \thead{Case 1} & \thead{Case 2} & \thead{Case 3} \\
		\hline
		\endfirsthead
		
		\cmidrule{1-5}\morecmidrules\cmidrule{1-5}
		\multirow{2}{*}{\thead{\quad Minimal \\ \quad Parameter}} & \multirow{2}{*}{\thead{True \\ Value}} & \multicolumn{3}{ c }{\thead{Estimated Value}} \\
		& & \thead{Case 1} & \thead{Case 2} & \thead{Case 3} \\
		\hline
		\endhead
						
		\hline \\
		\endfoot 
		
		\cmidrule{1-5}\morecmidrules\cmidrule{1-5}
		\endlastfoot
												
		$\hat{\pmb{\phi}}_{m}^{(1)}$ [Kg-$\text{m}^2$] & 2130.560 & 2126.814 & 1729.274 & 2121.443 \\ 
$\hat{\pmb{\phi}}_{m}^{(2)}$ [Kg-$\text{m}^2$] & 2078.560 & 2071.702 & 1930.767 & 2109.644 \\ 
$\hat{\pmb{\phi}}_{m}^{(3)}$ [Kg-$\text{m}^2$] & 1555.700 & 1554.130 & 1801.419 & 1740.076 \\ 
$\hat{\pmb{\phi}}_{m}^{(4)}$ [Kg-$\text{m}^2$] & -96.630 & -95.304 & -875.382 & -83.209 \\ 
$\hat{\pmb{\phi}}_{m}^{(5)}$ [Kg-$\text{m}^2$] & -269.450 & -267.023 & -607.177 & -294.371 \\ 
$\hat{\pmb{\phi}}_{m}^{(6)}$ [Kg-$\text{m}^2$] & -148.890 & -142.312 & 26.411 & -103.739 \\ 
$\hat{\pmb{\phi}}_{m}^{(7)}$ [Kg] & 2265.000 & 2244.299 & 1418.330 & 2198.283 \\ 
$\hat{\pmb{\phi}}_{m}^{(8)}$ [Kg-m] & 440.500 & 436.558 & 800.561 & 440.037 \\ 
$\hat{\pmb{\phi}}_{m}^{(9)}$ [Kg-m] & 679.500 & 673.810 & 574.592 & 658.626 \\ 
$\hat{\pmb{\phi}}_{m}^{(10)}$ [Kg-m] & 1033.000 & 1023.873 & 621.876 & 1005.331 \\ 
$\hat{\pmb{\phi}}_{m}^{(11)}$ [Kg-$\text{m}^2$] & -2.525 & -6.919 & 1305.865 & -34.097 \\ 
$\hat{\pmb{\phi}}_{m}^{(12)}$ [Kg-$\text{m}^2$] & 192.775 & 188.387 & 909.389 & -14.112 \\ 
$\hat{\pmb{\phi}}_{m}^{(13)}$ [Kg-$\text{m}^2$] & 7.400 & 10.166 & 614.689 & 4.942 \\ 
$\hat{\pmb{\phi}}_{m}^{(14)}$ [Kg-$\text{m}^2$] & 3.825 & 7.047 & 537.259 & 85.749 \\ 
$\hat{\pmb{\phi}}_{m}^{(15)}$ [Kg-$\text{m}^2$] & 6.000 & 2.818 & -521.054 & 0.024 \\ 
$\hat{\pmb{\phi}}_{m}^{(16)}$ [Kg-m] & 100.000 & 98.937 & -555.623 & 90.001 \\ 
$\hat{\pmb{\phi}}_{m}^{(17)}$ [Kg-m] & -4.500 & -4.993 & 253.047 & 5.614 \\ 
$\hat{\pmb{\phi}}_{m}^{(18)}$ [Kg-$\text{m}^2$] & -36.866 & -30.258 & -112.659 & -30.894 \\ 
$\hat{\pmb{\phi}}_{m}^{(19)}$ [Kg-$\text{m}^2$] & 55.134 & 56.976 & 240.184 & 62.475 \\ 
$\hat{\pmb{\phi}}_{m}^{(20)}$ [Kg-$\text{m}^2$] & 1.250 & 1.915 & 2.599 & 3.191 \\ 
$\hat{\pmb{\phi}}_{m}^{(21)}$ [Kg-$\text{m}^2$] & 1.670 & 1.422 & 129.586 & 6.837 \\ 
$\hat{\pmb{\phi}}_{m}^{(22)}$ [Kg-$\text{m}^2$] & -6.700 & -6.544 & 103.711 & -12.255 \\ 
$\hat{\pmb{\phi}}_{m}^{(23)}$ [Kg-m] & 46.000 & 45.310 & 74.935 & 45.001 \\ 
$\hat{\pmb{\phi}}_{m}^{(24)}$ [Kg-m] & -1.600 & -1.438 & -47.935 & -4.119 \\ 
$\hat{\pmb{\phi}}_{m}^{(25)}$ [Kg-$\text{m}^2$] & -14.010 & -11.737 & -28.911 & -12.295 \\ 
$\hat{\pmb{\phi}}_{m}^{(26)}$ [Kg-$\text{m}^2$] & 55.030 & 54.600 & 69.948 & 52.819 \\ 
$\hat{\pmb{\phi}}_{m}^{(27)}$ [Kg-$\text{m}^2$] & 1.380 & -0.975 & -15.320 & -1.341 \\ 
$\hat{\pmb{\phi}}_{m}^{(28)}$ [Kg-$\text{m}^2$] & 5.500 & 5.833 & -4.728 & 5.395 \\ 
$\hat{\pmb{\phi}}_{m}^{(29)}$ [Kg-$\text{m}^2$] & 3.930 & 3.865 & 2.534 & 2.835 \\ 
$\hat{\pmb{\phi}}_{m}^{(30)}$ [Kg-m] & 21.000 & 21.080 & 13.574 & 20.316 \\ 
$\hat{\pmb{\phi}}_{m}^{(31)}$ [Kg-m] & 12.000 & 12.092 & 7.293 & 12.556 \\ 
$\hat{\pmb{\phi}}_{m}^{(32)}$ [Kg-$\text{m}^2$] & 19.600 & 19.097 & 84.010 & 12.528 \\ 
$\hat{\pmb{\phi}}_{m}^{(33)}$ [Kg-$\text{m}^2$] & 256.670 & 258.902 & 463.905 & 302.331 \\ 
$\hat{\pmb{\phi}}_{m}^{(34)}$ [Kg-$\text{m}^2$] & -18.850 & -18.251 & 40.214 & -19.746 \\ 
$\hat{\pmb{\phi}}_{m}^{(35)}$ [Kg-$\text{m}^2$] & 2.330 & 2.474 & -477.795 & 11.882 \\ 
$\hat{\pmb{\phi}}_{m}^{(36)}$ [Kg-$\text{m}^2$] & 4.810 & 3.999 & 65.519 & -8.016 \\ 
$\hat{\pmb{\phi}}_{m}^{(37)}$ [Kg-m] & 122.500 & 121.080 & -12.235 & 115.634 \\ 
$\hat{\pmb{\phi}}_{m}^{(38)}$ [Kg-m] & 21.250 & 20.857 & -179.302 & 19.762 \\ 
$\hat{\pmb{\phi}}_{m}^{(39)}$ [Kg-$\text{m}^2$] & -66.290 & -67.589 & -103.468 & -74.876 \\ 
$\hat{\pmb{\phi}}_{m}^{(40)}$ [Kg-$\text{m}^2$] & 81.455 & 82.689 & -387.648 & 81.355 \\ 
$\hat{\pmb{\phi}}_{m}^{(41)}$ [Kg-$\text{m}^2$] & 2.112 & 1.306 & -12.430 & 2.533 \\ 
$\hat{\pmb{\phi}}_{m}^{(42)}$ [Kg-$\text{m}^2$] & 1.462 & 2.565 & 44.779 & 3.472 \\ 
$\hat{\pmb{\phi}}_{m}^{(43)}$ [Kg-$\text{m}^2$] & 21.483 & 20.749 & 206.239 & 20.474 \\ 
$\hat{\pmb{\phi}}_{m}^{(44)}$ [Kg-m] & 75.750 & 75.350 & -25.984 & 74.843 \\ 
$\hat{\pmb{\phi}}_{m}^{(45)}$ [Kg-m] & 1.750 & 1.832 & 54.533 & 2.982 \\ 
$\hat{\pmb{\phi}}_{m}^{(46)}$ [Kg-$\text{m}^2$] & -25.810 & -25.265 & -55.123 & -21.904 \\ 
$\hat{\pmb{\phi}}_{m}^{(47)}$ [Kg-$\text{m}^2$] & 60.370 & 61.114 & 84.505 & 61.583 \\ 
$\hat{\pmb{\phi}}_{m}^{(48)}$ [Kg-$\text{m}^2$] & 27.100 & 27.657 & 38.051 & 27.333 \\ 
$\hat{\pmb{\phi}}_{m}^{(49)}$ [Kg-$\text{m}^2$] & -1.980 & -2.074 & -5.270 & -1.736 \\ 
$\hat{\pmb{\phi}}_{m}^{(50)}$ [Kg-$\text{m}^2$] & 21.270 & 21.145 & 24.745 & 21.551 \\ 
$\hat{\pmb{\phi}}_{m}^{(51)}$ [Kg-m] & 36.000 & 35.662 & 21.960 & 35.448 \\ 
$\hat{\pmb{\phi}}_{m}^{(52)}$ [Kg-m] & -30.000 & -29.627 & -17.515 & -28.990 \\ 

	\end{longtable}
\end{center}

Further, we study the effect of the number of significant singular values of the regressor matrix on minimal parameter estimates. For this purpose, two trajectories with only the first 20 and 40 intervals are considered to estimate the parameters. The minimal parameter estimates are tabulated for 20 and 40 intervals under Case 2 and Case 3 respectively in Table \ref{min_params}. Median and maximum relative error in the parameters is calculated and tabulated in Table \ref{param_error}. 

The trajectory in Case 2 and Case 3 excites 38 and 46 out of the 52 minimal parameters respectively. In comparison to Case 1, the median and maximum relative errors are substantially higher for both the Case 2 and Case 3 because all the minimal parameters are not excited sufficiently. Although 46 out of 52 minimal parameters are excited in the trajectory with 40 intervals, the relative error is much higher compared to Case 1 reinforcing the idea that every parameter has to be sufficiently excited by the trajectory. Such an excitation is thoroughly obtained by the proposed framework at a much lower computational effort even with a bad initial guess.

\begin{table}[h!]
	\caption{Relative error in the estimated minimal parameters wrt. their true values}
	\label{param_error}
	\begin{center}
			\begin{tabular}{cccc}
				\cmidrule{1-4}\morecmidrules\cmidrule{1-4}
				& \thead{Case 1} & \thead{Case 2} & \thead{Case 3}\\
				\hline
				$\epsilon_{meadian}$ & 2.08\% & 150.22\% & 12.28\%\\
				$\epsilon_{max}$ & 174.02\% & 51817.41\% & 2141.79\%\\
				\cmidrule{1-4}\morecmidrules\cmidrule{1-4}
			\end{tabular}
	\end{center}
\end{table}

\subsection{Performance Evaluation of the Proposed Framework}
\begin{figure}[h!]
	\centering
	\includegraphics[width=0.7\linewidth, keepaspectratio]{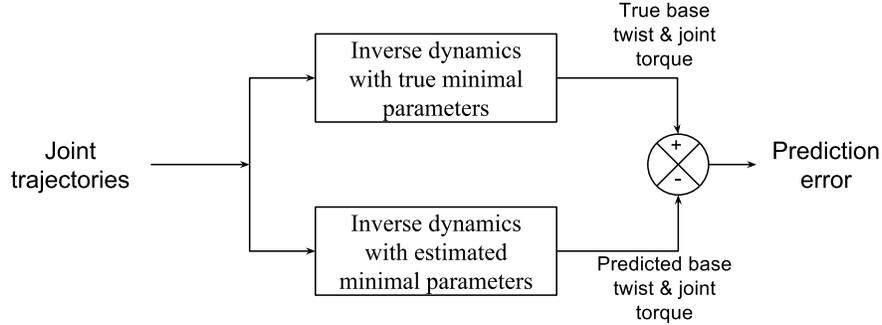}
	\caption{Block diagram of the performance evaluation scheme. The joint trajectory inputs to the inverse dynamics blocks are in Figure \ref{validation_traj}, and the prediction error outputs are in Figure \ref{state_prediction_error}.} 
	\label{perf_eval_chart}
\end{figure}

The primary goal of estimating the minimal parameters is to facilitate modeling, motion planning, and control of a space robot. The motion planning and control algorithms rely on the kinematic and dynamic model of the system. So, we use the estimated parameters to evaluate the accuracy with which they predict the kinematics and dynamics of the system, as in \cite{xu2017orbit, olsen2002maximum}. We evaluate the performance of the proposed framework by conducting inverse dynamics simulations with the true and estimated minimal parameters and computing the error in base twist and joint torques as depicted in Figure \ref{perf_eval_chart}. To consolidate the error trajectory into a single parameter, its RMS is computed.

\begin{figure}[t!]
	\centering
	\includegraphics[width=0.5\linewidth, keepaspectratio]{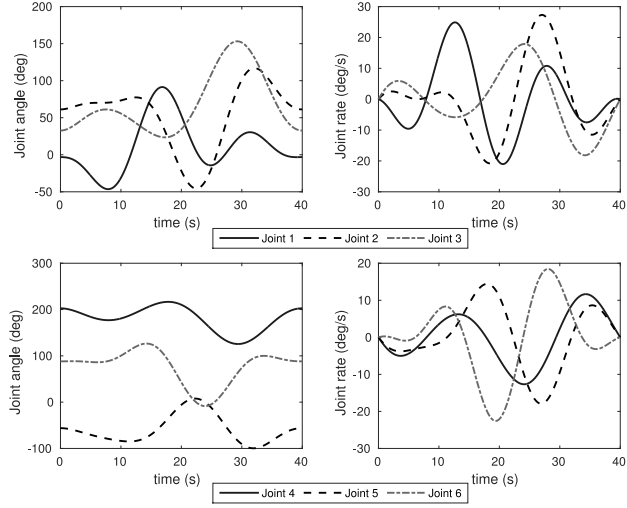}
	\caption{Fourier series based parametrization of joint angle and velocity trajectories used for performance evaluation of the estimated parameters} 
	\label{validation_traj}
\end{figure}

Cycloidal parametrization is used for parameter estimation. Hence, for performance evaluation of the estimated parameters, we consider a fourier series parametrization for joint position as follows:
\begin{flalign} \label{fourier_traj}
& q_{j} = q_{j, 0} + \sum_{i = 1}^{N_{h}}q^{(s)}_{j, i}\text{sin}(i\omega_{f} t) + q^{(c)}_{j, i}\text{cos}(i\omega_{f} t) \\
& \dot{q}_{j} = \sum_{i = 1}^{N_{h}}i\omega_{f}\big(q^{(s)}_{j, i}\text{cos}(i\omega_{f} t) - q^{(c)}_{j, i}\text{sin}(i\omega_{f} t)\big)
\end{flalign}
Here, $N_{h} = 2$ and $\omega_{f} = \pi/20$. The fourier based trajectories for joint angles and their rates are shown in Figure \ref{validation_traj}. 

\begin{figure}[t!]
	\centering
	\includegraphics[width=0.5\linewidth, keepaspectratio]{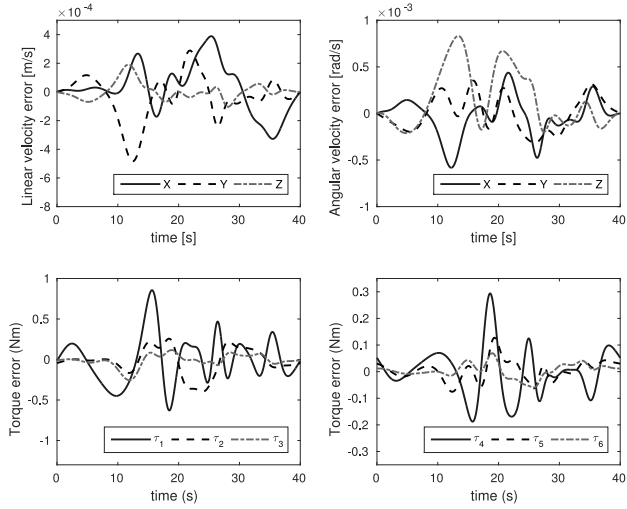}
	\caption{Error in the predicted base twist components and joint torques} 
	\label{state_prediction_error}
\end{figure}

The RMS of error for an arbitrary trajectory ($\psi$) is defined as follows:
\begin{flalign}
& \epsilon(\psi) = \sqrt{\frac{1}{N_{m}}\sum_{i=1}^{N_{m}}(\psi_{i} - \hat{\psi}_{i})^2} \label{rms_error}
\end{flalign}
We apply Eq. (\ref{rms_error}) to base twist and joint torque whose results are in Table \ref{perf_param}. The error trajectories are shown in Figure \ref{state_prediction_error}.

From Table \ref{perf_param}, it is clear that the RMS error in the predicted linear velocity is only 1 - 3 times higher than the standard deviation of its measurement noise (5e-5 m/s) and that of angular velocity is only 1 - 4 times higher than the standard deviation of its measurement noise (8e-5 rad/s). Further, the RMS error in predicted torque is much lower, i.e., by 1 - 10 times the standard deviation of the usual measurement noise of joint torque (0.3808 Nm \cite{olsen2002maximum}). With such low error in predicting the kinematics and dynamics of the system, it is safe to conclude that the parameters estimated using the proposed identification framework facilitates accurate analysis, motion planning, and control of a space robot.

\begin{table}
	\caption{RMS error in predicted base velocities and joint torques}
	\label{perf_param}
	\begin{center}
				\begin{tabular}{ccccccc}
					\cmidrule{1-7}\morecmidrules\cmidrule{1-7}
					\multirow{2}{*}{\thead{Base twist \\ component}} & $\textbf{\textit{v}}_{0}^{(x)}$ & $\textbf{\textit{v}}_{0}^{(y)}$ & $\textbf{\textit{v}}_{0}^{(z)}$ & $\pmb{\omega}_{0}^{(x)}$ & $\pmb{\omega}_{0}^{(y)}$ & $\pmb{\omega}_{0}^{(z)}$ \\ 
					& m/s & m/s & m/s & rad/s & rad/s & rad/s \\
					RMS error & 1.7e-4 & 1.6e-4 & 0.6e-4 & 2.1e-4 & 1.8e-4 & 3.5e-4 \\ \cmidrule{1-7}
					\multirow{2}{*}{\thead{Joint \\ torque}} & $\tau_{1}$ & $\tau_{2}$ & $\tau_{3}$ & $\tau_{4}$ & $\tau_{5}$ & $\tau_{6}$ \\
					& Nm & Nm & Nm & Nm & Nm & Nm \\
					RMS error & 0.2961 & 0.1665 & 0.0778 & 0.0922 & 0.0368 & 0.0262 \\
					\cmidrule{1-7}\morecmidrules\cmidrule{1-7}
				\end{tabular}
	\end{center}
\end{table} 

\section{Conclusion}
The applicability of the momentum model of tree-type space robots for minimal parameter identification is examined in this work.  Momentum model is linearly formulated in terms of the minimal parameters of a space robot. It is achieved by reformulating the generic momentum model and imposing the kinematic constraints posed by the space robot's joints. The minimal parameters uniquely model the space robot's momentum and dynamic models; hence, they are sufficient for kinematic and dynamic analysis, motion planning, and control. 

A systematic framework for computing the exciting trajectories is proposed based on an interval-wise approach. Each interval is constructed using a unique direction combination of the joints' velocity. A single interval excites a subset of minimal parameters; all the intervals excite the entire set of minimal parameters. The proposed framework is computationally efficient with only $2n$ design parameters for optimization and also results in only linear and bound constraints on the design parameters to adhere to the joint limits. Hence, the proposed framework is scalable with increasing degrees-of-freedom. Even though an initial guess of inertial parameters is required for computing exciting trajectories, it is shown that even a bad guess computed exciting trajctories that resulted in accurate parameter estimates.

Further, the minimal parameters of a 12 degrees-of-freedom, spatial, dual-arm space robot are identified with the kinematic data corrupted by noise. The estimated parameters predicted the base motions and joint torques accurately with prediction errors mostly in the order of noise in their measurements. Such accurate predictions with the proposed methodology even in the presence of noise facilitates accurate motion planning and control. The presented approach is generic and applicable to floating-base, tree-type robotic systems with reaction wheels. Consequently, it is also applicable to the underwater robots under the conservation of momentum.

\section*{Appendix}
\subsection{Analytical Expression of Minimal Parameters} \label{Appen:A}
The components of minimal parameter vector for the space robot shown in Figure \ref{sim_model} are as follows:
\begin{flalign*}
	& \pmb{\phi}_{m}^{(1)} = 
	\leftidx{^0}{\textbf{\textit{I}}}_{0}^{(xx)} + 	\leftidx{^1}{\textbf{\textit{I}}}_{1}^{(yy)} + \leftidx{^4}{\textbf{\textit{I}}}_{4}^{(yy)} + 0.9m_{1} + 1.9m_{2} + 1.9m_{3}  & \\
	& + 0.9m_{4} + 1.9m_{5} + 1.9m_{6} + 1.8(m_{1}\leftidx{^1}{\textbf{\textit{a}}}_{1}^{(z)} + m_{4}\leftidx{^4}{\textbf{\textit{a}}}_{4}^{(z)}) & \\
	& \pmb{\phi}_{m}^{(2)} = 
	\leftidx{^0}{\textbf{\textit{I}}}_{0}^{(yy)} + 	\leftidx{^1}{\textbf{\textit{I}}}_{1}^{(yy)} + \leftidx{^4}{\textbf{\textit{I}}}_{4}^{(yy)} + 1.3m_{1} + 2.3m_{2} + 2.3m_{3}  & \\
	& + 0.9m_{4} + 1.9m_{5} + 1.9m_{6} + 1.8(m_{1}\leftidx{^1}{\textbf{\textit{a}}}_{1}^{(z)} + m_{4}\leftidx{^4}{\textbf{\textit{a}}}_{4}^{(z)}) & \\
	& \pmb{\phi}_{m}^{(3)} = \leftidx{^0}{\textbf{\textit{I}}}_{0}^{(zz)} + 0.58(m_{1} + m_{2} + m_{3}) + 0.18(m_{4} + m_{5} + m_{6}) & \\
	& \pmb{\phi}_{m}^{(4)} = \leftidx{^0}{\textbf{\textit{I}}}_{0}^{(xy)} - 0.21(m_{1} + m_{2} + m_{3}) + 0.9(m_{4} + m_{5} + m_{6}) & \\
	& \pmb{\phi}_{m}^{(5)} = \leftidx{^0}{\textbf{\textit{I}}}_{0}^{(yz)} - 0.27(M - m_{0}) - 0.3(m_{1}\leftidx{^1}{\textbf{\textit{a}}}_{1}^{(z)} + m_{4}\leftidx{^4}{\textbf{\textit{a}}}_{4}^{(z)}) & \\
	& \pmb{\phi}_{m}^{(6)} = \leftidx{^0}{\textbf{\textit{I}}}_{0}^{(zx)} - 0.63(m_{1} + m_{2} + m_{3}) + 0.27(m_{4} + m_{5} + m_{6}) & \\
	& - 0.7m_{1}\leftidx{^1}{\textbf{\textit{a}}}_{1}^{(z)} + 0.3m_{4}\leftidx{^4}{\textbf{\textit{a}}}_{4}^{(z)} & \\
	& \pmb{\phi}_{m}^{(7)} = M & \\
	& \pmb{\phi}_{m}^{(8)} = m_{0}\leftidx{^0}{\textbf{\textit{a}}}_{0}^{(x)} + 0.7(m_{1} + m_{2} + m_{3}) - 0.3(m_{4} + m_{5} + m_{6}) & \\
	& \pmb{\phi}_{m}^{(9)} = m_{0}\leftidx{^0}{\textbf{\textit{a}}}_{0}^{(y)} + 0.3(M - m_{0}) & \\
	& \pmb{\phi}_{m}^{(10)} = m_{0} \leftidx{^0}{\textbf{\textit{a}}}_{0}^{(z)} + m_{1} \leftidx{^1}{\textbf{\textit{a}}}_{1}^{(z)} + m_{4} \leftidx{^4}{\textbf{\textit{a}}}_{4}^{(z)} + 0.9(M - m_{0})& \\
	& \pmb{\phi}_{m}^{(11)} = \leftidx{^1}{\textbf{\textit{I}}}_{1}^{(xx)} - \leftidx{^1}{\textbf{\textit{I}}}_{1}^{(yy)} + \leftidx{^2}{\textbf{\textit{I}}}_{2}^{(yy)} + \leftidx{^3}{\textbf{\textit{I}}}_{3}^{(yy)} - m_{2} & \\
	& \pmb{\phi}_{m}^{(12)} = \leftidx{^1}{\textbf{\textit{I}}}_{1}^{(zz)} + \leftidx{^2}{\textbf{\textit{I}}}_{2}^{(yy)} + \leftidx{^3}{\textbf{\textit{I}}}_{3}^{(yy)} + m_{2} + 2m_{3} & \\
	& \pmb{\phi}_{m}^{(13)} = \leftidx{^1}{\textbf{\textit{I}}}_{1}^{(xy)} + m_{2}\leftidx{^2}{\textbf{\textit{a}}}_{2}^{(z)} + m_{3}\leftidx{^3}{\textbf{\textit{a}}}_{3}^{(z)} & \\
	& \pmb{\phi}_{m}^{(14)} = \leftidx{^1}{\textbf{\textit{I}}}_{1}^{(yz)} & \\
	& \pmb{\phi}_{m}^{(15)} = \leftidx{^1}{\textbf{\textit{I}}}_{1}^{(zx)} & \\
	& \pmb{\phi}_{m}^{(16)} = m_{1}\leftidx{^1}{\textbf{\textit{a}}}_{1}^{(x)} + m_{2} + m_{3} & \\
	& \pmb{\phi}_{m}^{(17)} = m_{1}\leftidx{^1}{\textbf{\textit{a}}}_{1}^{(y)} - m_{2}\leftidx{^2}{\textbf{\textit{a}}}_{2}^{(z)} - m_{3}\leftidx{^3}{\textbf{\textit{a}}}_{3}^{(z)} & \\
	& \pmb{\phi}_{m}^{(18)} = 
	\leftidx{^2}{\textbf{\textit{I}}}_{2}^{(xx)} - 	\leftidx{^2}{\textbf{\textit{I}}}_{2}^{(yy)} - m_{3} & \\
	& \pmb{\phi}_{m}^{(19)} = 
	\leftidx{^2}{\textbf{\textit{I}}}_{2}^{(zz)} + m_{3} & \\
	& \pmb{\phi}_{m}^{(20)} = 
	\leftidx{^2}{\textbf{\textit{I}}}_{2}^{(xy)} & \\	
	& \pmb{\phi}_{m}^{(21)} = 
	\leftidx{^2}{\textbf{\textit{I}}}_{2}^{(yz)} & \\
	& \pmb{\phi}_{m}^{(22)} = 
	\leftidx{^2}{\textbf{\textit{I}}}_{2}^{(zx)} - 	m_{3}\leftidx{^3}{\textbf{\textit{a}}}_{3}^{(z)} & \\	
	& \pmb{\phi}_{m}^{(23)} = 
	m_{2}\leftidx{^2}{\textbf{\textit{a}}}_{2}^{(x)} + m_{3} & \\	
	& \pmb{\phi}_{m}^{(24)} = 
	m_{2}\leftidx{^2}{\textbf{\textit{a}}}_{2}^{(y)} & \\
	& \pmb{\phi}_{m}^{(25)} = 
	\leftidx{^3}{\textbf{\textit{I}}}_{3}^{(xx)} - \leftidx{^3}{\textbf{\textit{I}}}_{3}^{(yy)}& \\
	& \pmb{\phi}_{m}^{(26)} = 
	\leftidx{^3}{\textbf{\textit{I}}}_{3}^{(zz)} & \\
	& \pmb{\phi}_{m}^{(27)} = 
	\leftidx{^3}{\textbf{\textit{I}}}_{3}^{(xy)} & \\
	& \pmb{\phi}_{m}^{(28)} = 
	\leftidx{^3}{\textbf{\textit{I}}}_{3}^{(yz)} & \\
	& \pmb{\phi}_{m}^{(29)} = 
	\leftidx{^3}{\textbf{\textit{I}}}_{3}^{(zx)} & \\
	& \pmb{\phi}_{m}^{(30)} = 
	m_{3}\leftidx{^3}{\textbf{\textit{a}}}_{3}^{(x)} & \\
	& \pmb{\phi}_{m}^{(31)} = 
	m_{3}\leftidx{^3}{\textbf{\textit{a}}}_{3}^{(y)} & \\
	& \pmb{\phi}_{m}^{(32)} = \leftidx{^4}{\textbf{\textit{I}}}_{4}^{(xx)} - \leftidx{^4}{\textbf{\textit{I}}}_{4}^{(yy)} + 
	\leftidx{^5}{\textbf{\textit{I}}}_{5}^{(yy)} +
	\leftidx{^6}{\textbf{\textit{I}}}_{6}^{(yy)} - m_{5} & \\
	& \pmb{\phi}_{m}^{(33)} = \leftidx{^4}{\textbf{\textit{I}}}_{4}^{(zz)} + \leftidx{^5}{\textbf{\textit{I}}}_{5}^{(yy)} + 
	\leftidx{^6}{\textbf{\textit{I}}}_{6}^{(yy)} + m_{5} + 2m_{6} & \\
	& \pmb{\phi}_{m}^{(34)} = \leftidx{^4}{\textbf{\textit{I}}}_{4}^{(xy)} + m_{5} \leftidx{^5}{\textbf{\textit{a}}}_{5}^{(z)} + m_{6} \leftidx{^6}{\textbf{\textit{a}}}_{6}^{(z)} & \\
	& \pmb{\phi}_{m}^{(35)} = \leftidx{^4}{\textbf{\textit{I}}}_{4}^{(yz)} & \\
	& \pmb{\phi}_{m}^{(36)} = \leftidx{^4}{\textbf{\textit{I}}}_{4}^{(zx)} & \\
	& \pmb{\phi}_{m}^{(37)} = m_{4}\leftidx{^4}{\textbf{\textit{a}}}_{4}^{(x)} + m_{5} + m_{6} & \\
	& \pmb{\phi}_{m}^{(38)} = m_{4}\leftidx{^4}{\textbf{\textit{a}}}_{4}^{(y)} - m_{5}\leftidx{^5}{\textbf{\textit{a}}}_{5}^{(z)} - m_{6}\leftidx{^6}{\textbf{\textit{a}}}_{6}^{(z)} & \\
	& \pmb{\phi}_{m}^{(39)} = \leftidx{^5}{\textbf{\textit{I}}}_{5}^{(xx)} - \leftidx{^5}{\textbf{\textit{I}}}_{5}^{(yy)} - m_{6} & \\
	& \pmb{\phi}_{m}^{(40)} = \leftidx{^5}{\textbf{\textit{I}}}_{5}^{(zz)} + m_{6} & \\
	& \pmb{\phi}_{m}^{(41)} = \leftidx{^5}{\textbf{\textit{I}}}_{5}^{(xy)} \\
	& \pmb{\phi}_{m}^{(42)} = \leftidx{^5}{\textbf{\textit{I}}}_{5}^{(yz)} & \\
	& \pmb{\phi}_{m}^{(43)} = \leftidx{^5}{\textbf{\textit{I}}}_{5}^{(zx)} - m_{6}\leftidx{^6}{\textbf{\textit{a}}}_{6}^{(z)}& \\
	& \pmb{\phi}_{m}^{(44)} = m_{5}\leftidx{^5}{\textbf{\textit{a}}}_{5}^{(x)} + m_{6} & \\
	& \pmb{\phi}_{m}^{(45)} = m_{5}\leftidx{^5}{\textbf{\textit{a}}}_{5}^{(y)} & \\
	& \pmb{\phi}_{m}^{(46)} = \leftidx{^6}{\textbf{\textit{I}}}_{6}^{(xx)} - \leftidx{^6}{\textbf{\textit{I}}}_{6}^{(yy)}& \\
	& \pmb{\phi}_{m}^{(47)} = \leftidx{^6}{\textbf{\textit{I}}}_{6}^{(zz)} & \\
	& \pmb{\phi}_{m}^{(48)} = \leftidx{^6}{\textbf{\textit{I}}}_{6}^{(xy)} & \\
	& \pmb{\phi}_{m}^{(49)} = \leftidx{^6}{\textbf{\textit{I}}}_{6}^{(yz)} & \\
	& \pmb{\phi}_{m}^{(50)} = \leftidx{^6}{\textbf{\textit{I}}}_{6}^{(zx)} & \\
	& \pmb{\phi}_{m}^{(51)} = m_{6}\leftidx{^6}{\textbf{\textit{a}}}_{6}^{(x)} & \\
	& \pmb{\phi}_{m}^{(52)} = m_{6}\leftidx{^6}{\textbf{\textit{a}}}_{6}^{(y)} & \\
	& \text{where } M = \sum_{i = 0}^{6}m_{i} &
\end{flalign*}

\subsection{Rotation Matrices}
The rotation matrix of the base link ($\leftidx{^I}{\textbf{\textit{R}}_{0}}$) is formulated using the Z-X-Y euler angle notation. The link frames are assigned to the links using the modified DH notation \cite{khalil1986new}. Hence, for a parent-child pair $i, j \text{ such that } \rho_j = i$, the rotation matrix is: 
\begin{equation}
	\leftidx{^i}{\textbf{\textit{R}}_{j}}
	=
	\begin{bmatrix}
		Cq_{j} & -Sq_{j} & 0 \\
		Sq_{j} & Cq_{j}C \alpha_{j} & -S \alpha_{j} \\
		Sq_{j}S \alpha_{j} & Cq_{j}S \alpha_{j} & C \alpha_{j}
	\end{bmatrix}
\end{equation}
where, $C \bullet = cos(\bullet)$, $S \bullet = sin(\bullet)$.

\subsection{Computed Kinematic Data}
Computing the regressor matrix requires the kinematic data, i.e., base pose, base twist, joint angle, and joint velocity. However, only the rates are assumed to be measured and the pose is computed from the rates. Further, the base angular velocity in the inertial frame is computed from the euler rates. The quantities to be computed and measured is a choice that can be made by the user based on the available sensors. Finally, all the kinematic data is required to construct the regressor matrix. An time index `$t$' is appended to the subscript of the kinematic data. 

Base position and orientation at time index `$t$' are as follows:
\begin{align}
	\textbf{\textit{r}}_{0, t} = \textbf{\textit{r}}_{0, t - 1} + \textbf{\textit{v}}_{0, t-1}\Delta t \label{base_position} \\
	\pmb{\zeta}_{t} = \pmb{\zeta}_{t - 1} + \dot{\pmb{\zeta}}_{t-1}\Delta t \label{base_orientation}
\end{align}
where $\pmb{\zeta}$ is a 3 $\times$ 1 vector with euler angles.

Base angular velocity at time index `$t$' is as follows:
\begin{equation}
	\pmb{\omega}_{0, t} = \leftidx{^I}{\textbf{\textit{R}}_{0, t}} \textbf{\textit{M}}_{t}
	\dot{\pmb{\zeta}}_{t} \label{base_ang_vel}
\end{equation}

where, $\textbf{\textit{M}}_{t}$ transforms the euler rates to body rates.

Eq. \ref{base_position} and \ref{base_orientation} togther provide the base pose. Measured base velocity and the computed angular velocity in Eq. \ref{base_ang_vel} provide the base twist.

Joint angle of $j^{th}$ joint at time index `$t$' is computed from its joint velocity as follows:
\begin{equation}
	q_{j, t} = q_{j, t - 1} + \dot{q}_{j, t - 1} \Delta t \label{joint_angle}
\end{equation}

\section*{Acknowledgment}
This research was supported by Department of Science and Technology (DST), India INSPIRE research Grant (IFA-13 ENG-52) of the second author. The authors would like to thank J. Krishna Murthy, Robotics Research Center, International Institute of Information Technology, Hyderabad for his helpful suggestions on the early version of this paper.

\bibliography{references.bib}

\begin{thebibliography}{34}
\newcommand{\enquote}[1]{``#1''}
\providecommand{\natexlab}[1]{#1}
\providecommand{\url}[1]{\texttt{#1}}
\providecommand{\urlprefix}{URL }
\expandafter\ifx\csname urlstyle\endcsname\relax
  \providecommand{\doi}[1]{doi:\discretionary{}{}{}#1}\else
  \providecommand{\doi}{doi:\discretionary{}{}{}\begingroup
  \urlstyle{rm}\Url}\fi

\bibitem[{Flores-Abad et~al.(2014)Flores-Abad, Ma, Pham, and
  Ulrich}]{flores2014review}
Flores-Abad, A., Ma, O., Pham, K., and Ulrich, S., \enquote{A review of space
  robotics technologies for on-orbit servicing,} \emph{Progress in Aerospace
  Sciences}, Vol.~68, No. Supplement C, 2014, pp. 1 -- 26.
\newblock \doi{10.1016/j.paerosci.2014.03.002}.

\bibitem[{Umetani and Yoshida(1989)}]{umetani1989resolved}
Umetani, Y., and Yoshida, K., \enquote{Resolved motion rate control of space
  manipulators with generalized Jacobian matrix,} \emph{IEEE Transactions on
  Robotics and Automation}, Vol.~5, No.~3, 1989, pp. 303--314.
\newblock \doi{10.1109/70.34766}.

\bibitem[{Papadopoulos and Dubowsky(1991)}]{papadopolous1991nature}
Papadopoulos, E., and Dubowsky, S., \enquote{On the nature of control
  algorithms for free-floating space manipulators,} \emph{IEEE Transactions on
  Robotics and Automation}, Vol.~7, No.~6, 1991, pp. 750--758.
\newblock \doi{10.1109/70.105384}.

\bibitem[{Nanos and Papadopoulos(2015)}]{nanos2015avoiding}
Nanos, K., and Papadopoulos, E., \enquote{Avoiding dynamic singularities in
  Cartesian motions of free-floating manipulators,} \emph{IEEE Transactions on
  Aerospace and Electronic Systems}, Vol.~51, No.~3, 2015, pp. 2305--2318.
\newblock \doi{10.1109/TAES.2015.140343}.

\bibitem[{James et~al.(2016)James, Shah, Singh, Krishna, and
  Misra}]{james2016reactionless}
James, F., Shah, S.~V., Singh, A.~K., Krishna, K.~M., and Misra, A.~K.,
  \enquote{Reactionless Maneuvering of a Space Robot in Precapture Phase,}
  \emph{Journal of Guidance, Control, and Dynamics}, Vol.~39, No.~10, 2016, pp.
  2419--2425.
\newblock \doi{10.2514/1.G001828}.

\bibitem[{Peterson(2004)}]{peterson2004mass}
Peterson, W.~L., \enquote{Mass Properties Measurement in the X-38 Project,}
  \emph{63rd Annual Conference of the Society of Allied Weight Engineers},
  Paper 3325, Newport Beach, California, 2004.

\bibitem[{Gobbi et~al.(2011)Gobbi, Mastinu, and Previati}]{gobbi2011method}
Gobbi, M., Mastinu, G., and Previati, G., \enquote{A method for measuring the
  inertia properties of rigid bodies,} \emph{Mechanical Systems and Signal
  Processing}, Vol.~25, No.~1, 2011, pp. 305--318.
\newblock \doi{10.1016/j.ymssp.2010.09.004}.

\bibitem[{Schwartz and Hall(2004)}]{schwartz2004system}
Schwartz, J.~L., and Hall, C.~D., \enquote{System Identification of a Spherical
  Air-Bearing Spacecraft Simulator,} \emph{in Proceedings of the AAS/AIAA Space
  Flight Mechanics Conference, no. AAS 04-122}, 2004.

\bibitem[{James et~al.(2015)James, Vyas, Bandikatla, Mithun, and
  Shah}]{james2015design}
James, F., Vyas, S., Bandikatla, P., Mithun, P., and Shah, S.~V.,
  \enquote{Design and Development of an Earth Based Experimental Setup for
  Testing Algorithms on Space Robots,} \emph{Proceedings of the 2015 Conference
  on Advances In Robotics}, ACM, New York, USA, 2015, pp. 38:1--38:6.
\newblock \doi{10.1145/2783449.2783487}.

\bibitem[{Murotsu et~al.(1994)Murotsu, Senda, Ozaki, and
  Tsujio}]{murotsu1994parameter}
Murotsu, Y., Senda, K., Ozaki, M., and Tsujio, S., \enquote{Parameter
  identification of unknown object handled by free-flying space robot,}
  \emph{Journal of Guidance, Control, and Dynamics}, Vol.~17, No.~3, 1994, pp.
  488--494.
\newblock \doi{10.2514/3.21225}.

\bibitem[{Rackl et~al.(2013)Rackl, Lampariello, and
  Albu-Schäffer}]{rackl2013parameter}
Rackl, W., Lampariello, R., and Albu-Schäffer, A., \enquote{Parameter
  Identification Methods for Free-Floating Space Robots with direct Torque
  Sensing,} \emph{IFAC Proceedings Volumes}, Vol.~46, No.~19, 2013, pp. 464 --
  469.
\newblock \doi{10.3182/20130902-5-DE-2040.00121}, 19th IFAC Symposium on
  Automatic Control in Aerospace.

\bibitem[{Mayeda et~al.(1990)Mayeda, Yoshida, and Osuka}]{mayeda1990base}
Mayeda, H., Yoshida, K., and Osuka, K., \enquote{Base parameters of manipulator
  dynamic models,} \emph{IEEE Transactions on Robotics and Automation}, Vol.~6,
  No.~3, 1990, pp. 312--321.
\newblock \doi{10.1109/70.56663}.

\bibitem[{Lin(1995)}]{lin1995minimal}
Lin, S.-K., \enquote{Minimal linear combinations of the inertia parameters of a
  manipulator,} \emph{IEEE Transactions on Robotics and Automation}, Vol.~11,
  No.~3, 1995, pp. 360--373.
\newblock \doi{10.1109/70.388778}.

\bibitem[{Ayusawa et~al.(2013)Ayusawa, Venture, and
  Nakamura}]{ayusawa2014identifiability}
Ayusawa, K., Venture, G., and Nakamura, Y., \enquote{Identifiability and
  identification of inertial parameters using the underactuated base-link
  dynamics for legged multibody systems,} \emph{The International Journal of
  Robotics Research}, Vol.~33, No.~3, 2013, pp. 446--468.
\newblock \doi{10.1177/0278364913495932}.

\bibitem[{Atkeson et~al.(1986)Atkeson, An, and
  Hollerbach}]{atkenson1986estimation}
Atkeson, C.~G., An, C.~H., and Hollerbach, J.~M., \enquote{Estimation of
  Inertial Parameters of Manipulator Loads and Links,} \emph{The International
  Journal of Robotics Research}, Vol.~5, No.~3, 1986, pp. 101--119.
\newblock \doi{10.1177/027836498600500306}.

\bibitem[{Yoshida and Abiko(2002)}]{yoshida2002inertia}
Yoshida, K., and Abiko, S., \emph{Inertia Parameter Identification for a
  Free-Flying Space Robot}, Guidance, Navigation, and Control and Co-located
  Conferences, American Institute of Aeronautics and Astronautics, 2002.
\newblock \doi{10.2514/6.2002-4568}.

\bibitem[{Ma et~al.(2008)Ma, Dang, and Pham}]{ma2008orbit}
Ma, O., Dang, H., and Pham, K., \enquote{On-Orbit Identification of Inertia
  Properties of Spacecraft Using a Robotic Arm,} \emph{Journal of Guidance,
  Control, and Dynamics}, Vol.~31, No.~6, 2008, pp. 1761--1771.
\newblock \doi{10.2514/1.35188}.

\bibitem[{Christidi-Loumpasefski et~al.(2017)Christidi-Loumpasefski, Nanos, and
  Papadopoulos}]{christidi2017parameter}
Christidi-Loumpasefski, O.~O., Nanos, K., and Papadopoulos, E., \enquote{On
  parameter estimation of space manipulator systems using the angular momentum
  conservation,} \emph{2017 IEEE International Conference on Robotics and
  Automation (ICRA)}, 2017, pp. 5453--5458.
\newblock \doi{10.1109/ICRA.2017.7989641}.

\bibitem[{Nguyen-Huynh and Sharf(2013)}]{nguyen2013adaptive}
Nguyen-Huynh, T.~C., and Sharf, I., \enquote{Adaptive Reactionless Motion and
  Parameter Identification in Postcapture of Space Debris,} \emph{Journal of
  Guidance, Control, and Dynamics}, Vol.~36, No.~2, 2013, pp. 404--414.
\newblock \doi{10.2514/1.57856}.

\bibitem[{Xu et~al.(2017)Xu, Hu, Zhang, and Liang}]{xu2017orbit}
Xu, W., Hu, Z., Zhang, Y., and Liang, B., \enquote{On-orbit identifying the
  inertia parameters of space robotic systems using simple equivalent
  dynamics,} \emph{Acta Astronautica}, Vol. 132, No. Supplement C, 2017, pp.
  131 -- 142.
\newblock \doi{10.1016/j.actaastro.2016.12.031}.

\bibitem[{Dimitrov(2006)}]{dmitrov2006dynamics}
Dimitrov, D.~N., \enquote{Dynamics and control of space manipulators during a
  satellite capturing operation,} Ph.D. thesis, Tohoku University, 2006.

\bibitem[{Venture et~al.(2009)Venture, Ayusawa, and
  Nakamura}]{venture2009numerical}
Venture, G., Ayusawa, K., and Nakamura, Y., \enquote{A numerical method for
  choosing motions with optimal excitation properties for identification of
  biped dynamics - An application to human,} \emph{2009 IEEE International
  Conference on Robotics and Automation}, 2009, pp. 1226--1231.
\newblock \doi{10.1109/ROBOT.2009.5152264}.

\bibitem[{Lampariello and Hirzinger(2005)}]{lampariello2005modeling}
Lampariello, R., and Hirzinger, G., \enquote{Modeling and Experimental Design
  for the On-Orbit Inertial Parameter Identification of Free-Flying Space
  Robots,} \emph{Proceedings of ASME}, 2005, pp. 881--890.
\newblock \doi{10.1115/DETC2005-85242}.

\bibitem[{Sekhavat et~al.(2009)Sekhavat, Karpenko, and Ross}]{sekhavat2009ukf}
Sekhavat, P., Karpenko, M., and Ross, I., \emph{UKF-Based Spacecraft Parameter
  Estimation Using Optimal Excitation}, Guidance, Navigation, and Control and
  Co-located Conferences, American Institute of Aeronautics and Astronautics,
  2009.
\newblock \doi{10.2514/6.2009-5786}.

\bibitem[{Tanygin and Williams(1997)}]{tanygin1997mass}
Tanygin, S., and Williams, T., \enquote{Mass Property Estimation Using Coasting
  Maneuvers,} \emph{Journal of Guidance, Control, and Dynamics}, Vol.~20,
  No.~4, 1997, pp. 625--632.
\newblock \doi{10.2514/2.4099}.

\bibitem[{Gautier and Khalil(1992)}]{gautier1992exciting}
Gautier, M., and Khalil, W., \enquote{Exciting Trajectories for the
  Identification of Base Inertial Parameters of Robots,} \emph{The
  International Journal of Robotics Research}, Vol.~11, No.~4, 1992, pp.
  362--375.
\newblock \doi{10.1177/027836499201100408}.

\bibitem[{Xu et~al.(1993)Xu, Shum, Lee, and Kanade}]{xu1993adaptive}
Xu, Y., Shum, H.-Y., Lee, J.-J., and Kanade, T., \enquote{Adaptive control of
  space robot system with an attitude controlled base,} \emph{Space Robotics:
  Dynamics and Control}, Springer, 1993, pp. 229--268.

\bibitem[{Khosla(1989)}]{khosla1989categorization}
Khosla, P.~K., \enquote{Categorization of parameters in the dynamic robot
  model,} \emph{IEEE Transactions on Robotics and Automation}, Vol.~5, No.~3,
  1989, pp. 261--268.
\newblock \doi{10.1109/70.34762}.

\bibitem[{Gautier(1991)}]{gautier1991numerical}
Gautier, M., \enquote{Numerical calculation of the base inertial parameters of
  robots,} \emph{Journal of Robotic Systems}, Vol.~8, No.~4, 1991, pp.
  485--506.
\newblock \doi{10.1002/rob.4620080405}.

\bibitem[{Gautier and Khalil(1990)}]{gautier1990direct}
Gautier, M., and Khalil, W., \enquote{Direct calculation of minimum set of
  inertial parameters of serial robots,} \emph{IEEE Transactions on Robotics
  and Automation}, Vol.~6, No.~3, 1990, pp. 368--373.
\newblock \doi{10.1109/70.56655}.

\bibitem[{Rackl et~al.(2012)Rackl, Lampariello, and Hirzinger}]{rackl2012robot}
Rackl, W., Lampariello, R., and Hirzinger, G., \enquote{Robot excitation
  trajectories for dynamic parameter estimation using optimized B-splines,}
  \emph{Robotics and Automation (ICRA), 2012 IEEE International Conference on},
  IEEE, 2012, pp. 2042--2047.
\newblock \doi{10.1109/ICRA.2012.6225279}.

\bibitem[{Shah et~al.(2012)Shah, Nandihal, and Saha}]{shah2012recursive}
Shah, S.~V., Nandihal, P.~V., and Saha, S.~K., \enquote{Recursive dynamics
  simulator (ReDySim): A multibody dynamics solver,} \emph{Theoretical and
  Applied Mechanics Letters}, Vol.~2, No.~6, 2012, p. 063011.
\newblock \doi{10.1063/2.1206311}.

\bibitem[{Olsen et~al.(2002)Olsen, Swevers, and Verdonck}]{olsen2002maximum}
Olsen, M.~M., Swevers, J., and Verdonck, W., \enquote{Maximum Likelihood
  Identification of a Dynamic Robot Model: Implementation Issues,} \emph{The
  International Journal of Robotics Research}, Vol.~21, No.~2, 2002, pp.
  89--96.
\newblock \doi{10.1177/027836402760475379}.

\bibitem[{Khalil and Kleinfinger(1986)}]{khalil1986new}
Khalil, W., and Kleinfinger, J., \enquote{A new geometric notation for open and
  closed-loop robots,} \emph{Proceedings. 1986 IEEE International Conference on
  Robotics and Automation}, Vol.~3, IEEE, 1986, pp. 1174--1179.
\newblock \doi{10.1109/ROBOT.1986.1087552}.

\end{thebibliography}
\end{document}